\documentclass[submit]{Definitions/mdpi} 
\renewcommand{\linenumbers}{}
\firstpage{1} 
\pubvolume{1}
\issuenum{1}
\articlenumber{0}
\pubyear{2025}
\copyrightyear{2025}
\datereceived{ } 
\daterevised{ } 
\dateaccepted{ } 
\datepublished{ } 
\hreflink{https://doi.org/} 

\usepackage{array}
\usepackage{amsthm}
\usepackage{enumitem}

\newcolumntype{Y}[1]{>{\raggedright\let\newline\\\arraybackslash\hspace{0pt}}p{#1}}
\theoremstyle{definition}
\newtheorem{defin}{Definition}

\Title{Co-creation for Sign Language Processing and Machine Translation}

\TitleCitation{Co-creation for Sign Language Processing and Machine Translation}

\Author{Lisa Lepp $^{1}$, Dimitar Shterionov $^{2}$\orcidA{}, Mirella De Sisto $^{3}$\orcidB{} and Grzegorz Chrupa{\l}a $^{4}$\orcidC{}}

\AuthorNames{Lisa Lepp, Dimitar Shterionov, Mirella De Sisto and Grzegorz Chrupa{\l}a}

\address{%
$^{1}$ \quad Lisa Lepp: l.b.lepp@tilburguniversity.edu\\
$^{2}$ \quad Dimitar Shterionov: d.shterionov@tilburguniversity.edu\\
$^{3}$ \quad Mirella De Sisto: m.desisto@tilburguniversity.edu\\
$^{4}$ \quad Grzegorz Chrupa{\l}a: grzegorz@chrupala.me}

\abstract{Sign language machine translation (SLMT) -- the task of automatically translating between sign and spoken or between sign and sign languages -- is a complex task within the field of NLP. Its multi-modal and non-linear nature require the joint efforts of sign language (SL) linguists, technical experts and SL users. Effective user involvement is a challenge that can be addressed through \emph{co-creation}. Co-creation has been formally defined in many fields, e.g. business, marketing, educational and others, however in NLP and in particular in SLMT there is no formal, widely accepted definition. Starting from the inception and evolution of \textit{co-creation} across various fields over time, we develop a relationship typology to address the collaboration between deaf, Hard of Hearing and hearing researchers and the co-creation with SL-users. We compare this new typology to the guiding principles of participatory design for NLP. We, then, assess 111 articles from the perspective of involvement of SL users and highlight the lack of involvement of the sign language community or users in decision-making processes required for effective co-creation. Finally, we derive formal guidelines for co-creation for SLMT which take the dynamic nature of co-creation throughout the life cycle of a research project into account.}

\keyword{Sign Language Machine Translation; Co-creation; Sign Language Community; Relationship typology; Techniques; Sign Languages; Machine Translation; }

\begin{document}

\section{Introduction}
In the last two decades Machine Translation (MT) has become an indispensable aid for reducing language barriers and facilitating quick exchange of and access to information. MT has found a dominant place in the workflow of (professional and non-professional) translators and translation services. Since its inception in the 1950s, MT has evolved rapidly reaching unprecedented qualities thanks to the advances in AI, deep learning and other related (sub)fields, the innovations of the core processing technology (GPUs instead of CPUs), the raising volumes and quality of data and others. These advances have had a significant impacted on text-to-text translation; however, when it comes to sign languages (SLs), used primarily by deaf and hard of hearing (HoH) individuals\footnote{Historically, it has been the case that that researchers use the distinctive categories such as 'hearing', 'HoH', or 'deaf', with lower letters instead of with capitals, such as Deaf, which is showing the cultural- and social gains of being deaf. As we try to place this article within a historical framework, we stick to the use of these words with lower capitals, while we acknowledge that this division should originally be between signers and non-signers. Within the group of signers you can have different sub-categories. Hearing does not imply that a person cannot sign, just as being deaf does not automatically means that they can sign. However, as we will see in the current paper, researchers without knowledge or skills in SLs are making decisions that are in contrast to what is needed or benefiting the sign language community (SLC).}, the advances are very limited. These limitations can be attributed to many technical factors, e.g. data, processing capacity, methodological gaps and others, but a large one is the human factor. Research on and with SLs, and especially in the context of developing SL technology (SLT), has not only been predominantly hearing-led, but also involving deaf people in an ineffective and often in an unethical way. While this trend is changing, there are many issues which need to be addressed in the process. Most of these issues are related to two main questions. First, how to \emph{establish a relationship} between the researcher / developer and the citizen / end-user/ community, which can allow the active, ethical and fair involvement of both parties into the ideation, design, production and/or delivery of research and development activities. Second, how to \emph{maintain such a relationship} in order for both parties (as well as society in general) to benefit. The latter relates to more granular decisions such as how to determine who owns the data, how to ensure privacy during data-collection and utilization, who decides what proper SL-use is, which kind of groups of SL-users need to be involved, how much diversity is included, what the effect or impact is of such kind of projects for the SLC, and in the end, who is really benefiting from such projects. 

In this article we address the aforementioned issues and formally define \emph{co-creation} for SL processing (SLP) and machine translation (SLMT) research and development. We adapt the participation relationship typology of~\cite{Harderetal2013} to the specifics of SLP and SLMT and align it with the guidelines for participatory research by~\citet{caselli-etal-2021-guiding}. The proposed definition sets a proper foundation to allow researchers and practitioners in the field to establish the right communication and collaboration protocols and engage in effective co-operation.

This article is structured as follows: in Section~\ref{sec:co-creation}, we first provide an overview of co-creation as defined and used in many different fields. We discuss the relationship between researcher and the end-user, focusing on the typology of~\cite{Harderetal2013} in Section~\ref{sec:relationship}. In Section~\ref{sec:co-creation-sLMT}, we discuss which adaptions are necessary for a proper usage of the existing typology for the SL-user; we then review 111 articles on SLMT and categorize them based on the new typology. We wrap up the article by proposing formal guidelines for co-creation in the context of linguistic (theoretical) research and (practical) development in the field of SLMT in Section~\ref{sec:formal guidelines}. 

\section{Co-creation}\label{sec:co-creation}
\emph{Co-creation}, a term that has gained significant attention recently, has been used in many fields -- including business, marketing, educations and others. \citeauthor{Prahalad2000co} used this term for the first time to point at the process of involving consumers in the creation of the project value, and their competing for value~\cite{Prahalad_co-creation_2004}. Co-creation evolved from \textit{cooperative design} -- a movement in the 1960s in Sweden and Denmark in which workers participated in the design of IT systems that impacted their work~\cite{Goransdotter2018-Scandinavia-User-centred-Design}. Later, in the 1970s, in the US similar methodology was adopted under the name \textit{participatory design} and in the 80s urban planners adopted the concept in the form of \textit{collaborative planning}. In 2010, co-creation was the focus of the book ``The Power of Co-creation'' by~\citet{ramaswamy2010power} -- a business-oriented manuscript, which describes that co-creation is about involving people in a collaborative process, wherein businesses and customers are working together to create a value, product, or experiences, which in turn boosts the overall value and the benefits of the (collaboration) network involved. In the current article we adopt the definition of co-creation pinned by~\citet{Frow2015_Co-design}: 

\begin{defin}
    Co-creation is a process which establishes and maintains a dialogue between users and developers not only in the research phase, but also in the phases of ideation, prototyping and implementation. 
\end{defin}

As with many other methodologies, co-creation is not an ``all or nothing'' phenomena (either you co-create or you do not), but multi-layered with varying levels of involvement of the users. In the context of SLMT research, some SLMT projects involve the SLC for evaluation of models~\cite{Almeidaetal2015, Chiuetal2006, Foongetal2009, Stein2006, Wuetal2007}, for data-collection~\cite{Guoetal2019, Koetal2019, Morrissey2007}, in experimental situations~\cite{López-Ludeñaetal2014, Mazzei2013, su_wu2009}, or in workshops~\cite{López-Ludeñaetal2014} and interviews~\cite{DavidBouillon2018}. Others attempt to cooperate with or integrate the SLC in their research~\cite{Baldassarrietal2009, Baueretal1999, MorrisseyWay2013, Sagawaetal1996}. However, even when users are involved, the size and diversity of the population is of importance and often the included population is not a representation of the (variation in between the) SLC. For example, in~\cite{Mistryinden2018}, it is not clear whether the participants were deaf, or in~\cite{Adnanetal2013} where the participants were not native signers.


To define a co-creation methodology for SLP and in particular SLMT we review works related to different forms of co-creation, as defined and used in various fields. We review works on participatory research, participatory or collaborative design (co-design) and co-creation. Later, we establish a co-creative approach to SLP research that is in line with existing works, fits the use-cases of SLP and can be positioned among these works.

\paragraph{\textbf{Marketing / Business perspective}} Traditionally, innovators, developers, businesses, etc., produced goods or services for customers to buy or use with little to no involvement of the target user groups during the development process. 
This obsolete approach evolved to allow iterative dialogues between producer and consumer that can create value for both parties~\cite{Ballantyne_dialogue_2004} and has been referred to as co-production in the business literature (see~\cite{bendapudi_psychological_2003} for a detailed overview of fundamental literature related to co-production) and later, as remarked in~\cite{vargo_service-dominant_2006}, as co-creation. As noted by~\citet{Vargo2004}, the latter term better aligns with the concept of service-dominant logic, a marketing framework suggested by~\citeauthor{Prahalad2000co}, which sees customers as actors who create and compete for value~\cite{Prahalad2000co}. Consumers as source of competence, bring their knowledge and skills, their willingness to learn and experiment, and engage in active dialogues. This competence, in addition, brings the consumer a competitive advantage, allowing them to judge and negotiate terms and prices, therefore, to extract value from enterprises~\cite{Prahalad2000co}.


\paragraph{\textbf{Social domain}} Within the social domain, co-creation refers to \textit{the collaboration between a variety of actors actively joining forces to tackle jointly defined challenges}~\cite{StierSmit2021}. Actors might belong to various sectors of the society, often spanning over academia, government, industry and societal partners.

\paragraph{\textbf{Community-based participatory research (CBPR)}}
The main goal of CBPR is to bring academics and community members into research partnerships \cite{Stack2018}. In several fields like healthcare, the public domain, social domain, and public sector, working together with the citizen or customer became the new, modern norm~\cite{Voorberg2014}. The goal of cooperation with the community in the public domain, for example, is to meet societal needs by fundamentally altering the relationships, roles, and rules among the stakeholders involved, through an open process of participation, exchange, and collaboration with relevant parties, including end-users, thus transcending organizational boundaries and jurisdictions\cite{Voorberg2014}.

\paragraph{\textbf{Responsible research and innovation (RRI)}} Science and innovation outputs resonate across direct and indirect stakeholders spreading over whole societies. To increase positive impacts and lower the negative ones, the concept of RRI implies a shift ``from solutions developed internally within the research community and only tolerated passively by society towards ones that are taking citizens and other actors actively into consideration as part of the development of solutions that are more apt to achieve desirable results with a high impact''~\cite{owen2019responsible}. 

\paragraph{\textbf{In NLP and MT}}
The importance of involving user communities is also quite relevant for the field of NLP. However, in the field of NLP, the term ``co-creation'' often is used in the context of human-AI collaborations for effective content creation. Such involve the recent work by~\citet{sharma2024-investigating-co-creation,konen-etal-2024-style-co-creation,ding-etal-2023-harnessing-co-creation}, among others, who investigate how large language models (LLMs) and (human) content creators can most effectively collaborate. Co-creative approaches have also been employed for specific tasks such as poetry generation~\cite{goncalo-oliveira-etal-2017-co-creation}, literature synthesis~\cite{manjavacas-etal-2017-synthetic-literature-co-creation}, interpreting~\cite{nakaguchi-etal-2016-combining-co-creation-for-interpreting} and others.\footnote{The ACL anthology search of 7-Oct-2024 results in 146 relevant works.}

Another form of co-creation in NLP is by involving users as knowledge resource for training models. However, to refer to this process as co-creation, following the definition of co-creation as an ongoing dialog, researchers should consider works that continuously involve users in the various model, tool or project development stages. 
\citet{caselli-etal-2021-guiding} highlight that too often language data is gathered in a way that does not consider nor involve language users. This is a broad problem in NLP, which is not only limited to language minorities but also affects well-represented languages. \citeauthor{caselli-etal-2021-guiding} call for shifting the perspective from \textit{language as data} to \textit{language as people}; given that language is produced by humans and has a deep function in human life. They propose 9 guiding principles inspired by participatory design. 
These principles underline that design is a continuous process, recognize the importance of communication with communities and of creating understanding and trust and aim to stir NLP towards producing what the communities need and not `replacing humans', including community feedback in the NLP pipeline. In addition, the work acknowledges that making one checklist that is valid for every community and every circumstances is impossible. These guidelines are presented in detail in Appendix~\ref{sec:app_guidelines}.

It is worth noting that participatory design has been adopted by the early work of~\cite{ciravegna-etal-1997-participatory-linguistic-engineering} which involves engineers and users in ``linguistic engineering'' process through a tool, called GEPPETTO, designed to support both roles. However, the embedded practices did not exceed the scope of the work of~\citet{ciravegna-etal-1997-participatory-linguistic-engineering}; parts of the developed tool were integrated in the work of~\cite{lavelli-etal-2001-sissa}.

\paragraph{Participatory research, co-design and co-creation}
While focusing on co-creation the above also covers definitions related to participatory research, participatory or collaborative design (or co-design) and co-creation. It is important to note that these are not the same and despite their overlaps we ought to acknowledge their differences. \textbf{Participatory research} is the process of involving users as domain experts at all levels of research process. \textbf{Participatory design} is about engaging with end users as experts in their individual experiences. The contrast between participatory research and participatory design (or co-design) is the purpose of involvement -- in the former end users aid the research and development process while in the latter, they contribute to identifying possible solutions. \textbf{Co-creation}, as defined above, is a form of engagement between businesses and end users that is also about identifying new and emergent value opportunities for both parties.\footnote{More details on the differences and similarities between these three terms can be found in Josh Morrow's post~\cite{morrow_co-creation_2022}}

\section{The essential elements of co-creation — a common goal and mutual relationship}\label{sec:relationship}
Section~\ref{sec:co-creation} presents \textit{different} definitions from \textit{different} fields and for \textit{different} purposes often revolving around several goals. This indicates that the formalization as well as application of co-creation are not universal but strongly depend on the involved stakeholders and use-cases. 

We identify two common threads: (i) a \textbf{shared / common goal} and (ii) a \textbf{formal and sustainable relationship} between end users or consumers and researchers, developers or businesses. The \textbf{shared goal} is the starting point of every cooperative research project as clearly pointed out by~\citet{StierSmit2021}. Each partner or stakeholder needs to define their problem to make sure that the objectives of all parties are aligned. Academics typically refer to an issue as a research problem, whereas a civil servant might describe it as a political or health problem \cite{StierSmit2021}. \citet{StierSmit2021} also stress the importance of being frank early in the collaboration process with what one sees as one's role in the co-creation. In doing so, it may be necessary to convey the objectives and underpinning values of one's work and organization. Unclear roles can place responsibilities on the wrong participants raising ethical dilemmas. Building a sustainable relationship with the citizen/end-user/actor/community, and trying to involve them actively into the design, production or delivery, e.g. not only as passive participant, but as an actively-contributing and valuable \textit{partner} with a \textit{joint responsibility}, is key to co-creation.


In~\cite{Harderetal2013}, \citeauthor{Harderetal2013} set up a framework of a participation relationship between researcher (A) and society (B) spread over three dimensions -- \textit{depth}, \textit{breadth} and \textit{scope} of participation -- organised in a typology of six levels. These capture the typical processes, attitudes, assumptions (of A) and actions (of A) towards the users or society (B).  
Table~\ref{tbl:typology} gives an overview of~\cite{Harderetal2013}'s typology of the relationships between A and B in terms of the typical processes.


\begin{table}[!h]
    \centering
    \begin{tabular}{|Y{1.85cm}|Y{1.85cm}|Y{1.85cm}|Y{1.85cm}|Y{1.85cm}|Y{1.85cm}|}\hline
         Level (-1) &  Level (0) & Level (1) & Level (2) & Level (3) & Level (4) \\\hline
         \textit{Denigration} &  \textit{Neglect} & \textit{Learning about} & \textit{Learning from} & \textit{Learning together} & \textit{Learning as one} \\\hline
         A makes decisions without B's involvement (sometimes contrary to B's interests). & A makes decisions without B's involvement: ignorant or dismissive of B's interests & A asks B's opinions, but does not feel obliged to take them into account: A makes the final decisions. & A asks B's opinions and considers B's contribution seriously. A still makes the final decision. & Major issues are negotiated through discussion between A and B. Most decisions are made jointly, e.g. by consensus-building. & A-B consortium discusses relevant issues by focusing on the ideas themselves, rather than the source of ideas.\\\hline
    \end{tabular}
    \caption{A typology of relationships of participation between researchers (A) and society (B).~\cite{Harderetal2013}}
    \label{tbl:typology}
\end{table}



In \textbf{Levels -1 (Denigration) and 0 (Neglect)} there is no involvement of the user community, only researchers are involved in development and decision-making. Society's interests or needs are not investigated nor taken into account. While level 0 covers various forms of negligence of the researchers towards the actual needs and wishes of the users, -1 points towards a disruptive attitude with negative impact. In \textbf{Level 1 (Learning About)} the knowledge of society is acknowledged, but there is no influence (of the society / user-community) on the developed product or the development process; \textbf{Level 2 (Learning From)} considers an active participation, but without influence on major decision points. Although this level number indicates positive processes to include the society, they are still on the lower part of the ladder, with substantial decision making \textit{power} in the hands of the researcher(s), where automatically the society is having less power by the fact that they cannot influence the decision-making process. 
According to \textbf{Level 3 (Learning Together)} discussions about relevant topics are held between A and B, there is consensus about the outcomes, and most decisions are jointly made. \textbf{Level 4 (Learning As One)} represents a \textit{full partnership} where each stakeholder is entirely involved, and their skills, knowledge and life-experience are valued and employed to succeed in a shared goal.\newline

    

\noindent In the context of these levels, different dimensions are clearly defined as:

\begin{itemize}
    \item \textit{Depth} refer to the \textit{extent of control over decision-making} by the involved participants. It also refers to the amount of \textit{power} that each party is having in the project. \citet{Harderetal2013} uses the term \textit{depth} for the first time in an educational setting, where they talk about lower and higher status of the stakeholders.
    \item \textit{Breadth} refers to the extent of \textit{diversity} that the groups covers. Who are the stakeholders? Participation can be divided into different groups, such as leader, wider society, advisers, technical teams and so on. The idea behind this division into several groups, is to include as much diversity as possible. However, this does not mean that participants from one category cannot ``join'' other groups. These boundaries need to be discussed with the society itself. 
    \item \textit{Scope} refers to the \textit{various stages} of the decision-making. This aspect of participation contains the initiation, the planning of the design, its implementation, the reflection, the communication, and the expected outcomes, not only on the short-term, but also the long-term ones. 
\end{itemize}


\section{Co-creation for SLMT research}\label{sec:co-creation-sLMT}
In this section we define co-creation for SLMT through (i) definition of a common goal; (ii) an adaptation of \citet{Harderetal2013}'s typology into a basic and advanced versions for SLC participation and (iii) embedding of the guidelines of~\cite{caselli-etal-2021-guiding}. 

\subsection{Common goal} 
As noted above, co-creation is a process which revolves around a common goal. We formulate a description of a high-level common goal as:

\begin{defin}
    The research and development of technological solutions to the understanding and processing of Sign Languages and the translation between Sign and Spoken languages (and by proxy between sign and sign languages) to 
    facilitate or improve the communication between hearing, HoH and deaf individuals as well as to improve the access and dissemination of multilingual, multi-modal (video, audio and text) information.
\end{defin}

\subsection{Basic relationship typology for SLC participation}\label{sec:new_typology_1}
The typology showed in Table~\ref{tbl:typology} formalities the relationships of participation between actors A and B \cite{Harderetal2013}. We explicitly address the involvement of SLCs placing deaf, HoH and hearing participants in the corresponding roles (A or B). 

Both groups of participants -- A and B -- can generally involve deaf, HoH and hearing individuals. Actor A -- as used by \cite{Harderetal2013} -- refers to the academics who conduct the research. In our case, these are professionals who conduct research in the field of SLMT, SLNLP, and SLP-technology, a role which predominantly has been assumed by \textbf{hearing} researchers as evident by the literature review in Section~\ref{sec:co-creation in reviewed articles}. Therefore, in our typology the role A is assumed by the \textit{hearing researcher}. As noted earlier, in general, A should involve all three categories of actors. That is, researchers (A) do not necessarily need to be hearing and that in an ideal situation deaf, HoH and hearing researchers work together. There is a (slow) ongoing shift, where more and more HoH and deaf researchers participate in SLMT and SLP research and, although rarely still, assume leading roles. Our typology positions these roles on a spectrum from negative to positive, with the division between hearing-led roles that contradict the perspectives of SLCs (level -1) and collaborative roles (level 4).

Actor B, in contrast, includes the SLC, the SL-user, and the HoH and deaf academic researchers. We ought to stress the following. First, not every SL-user is part of a SLC by default. This term might raise the idea that a SLC includes all variations in SL-fluency, equal access to a visual language, and educational possibilities. However, this is not the case as noted by the National Association of the Deaf~\cite{NAD_CommunityAndCulture}. Therefore, we make the distinction between SLC and SL-users so that even SL-users that are outside of an SLC, e.g. (Hearing) Children of Deaf Adults (CODA's), are regarded as well as other individuals which belong to an SLC. Second, the current typology does not clearly quantify the amount of \textit{power} distributed between hearing, and HoH or deaf researchers, as the focus is mainly on co-creation with the society. \citet{Harderetal2013} uses the term \textit{depth} for the first time in an educational setting, where they talk about lower and higher status of the stakeholders.
This is also valid for \emph{breadth} and \emph{scope}. However, it is worth noting that the distribution of power, diversity and involvement in different stages ranges from the one extreme point where all the power of decision making is in the hands of the hearing researchers who compose the complete consortium or collaboration network and are solely involved in all stages of the project, to a balance point where consortia are built from deaf and hearing researchers, who collaborate on equal terms, involve the SL-user in all stages of the project and all stakeholders can excerpt the power of decision making. 

The adapted typology is shown in Table~\ref{tbl:new_typology}.


\begin{table}[h]
    {\setlength{\tabcolsep}{2pt}
    \small
    \centering
    \begin{tabular}{|Y{2.1cm}|Y{2.0cm}|Y{2.1cm}|Y{2.1cm}|Y{2.3cm}|Y{2.1cm}|}\hline
         Level (-1) &  Level (0) & Level (1) & Level (2) & Level (3) & Level (4) \\\hline
         \textit{Denigration} &  \textit{Neglect} & \textit{Learning about} & \textit{Learning from} & \textit{Learning together} & \textit{Learning as one} \\\hline
         Hearing researchers make decisions without the SLC (neither HoH or deaf researchers) involvement, contrary to the SLCs interests. & Hearing researchers make decisions without the SLC (neither HoH or deaf researchers) involvement, ignorant or dismissive of the SLCs interests. & Hearing researchers ask the SLCs and the users (and/or HoH or deaf researchers) opinions, but do not necessarily take them into account: the hearing researchers make the final decisions. & Hearing researchers ask the SLCs and the users opinions and consider the SLCs and users seriously. Hearing researchers still makes the final decision based on the information, HoH and deaf researchers are asked for evaluation, but not included in the process. & Major objectives and issues are discussed / negotiated jointly involving hearing, HoH and deaf researchers, and SL-users. Most decisions are made jointly, e.g. by consensus-building. & A consortium that includes hearing, HoH and deaf researchers, and SLC members, jointly built, discuss relevant issues by having knowledge exchange (e.g. seminars on different topics from all involved communities).\\\hline
    \end{tabular}
    }
    \caption{A typology of relationships of participation between deaf, hearing and HoH participants -- both as users and as researchers / developers.}
    \label{tbl:new_typology}
\end{table}

\subsection{Advanced relationship typology for SLC participation} \label{advanced typology}
The 4th level as presented in Table~\ref{tbl:new_typology} does not capture the long-term potential for impact such a collaboration may have on the various participants. To align with principles 2, 3 and 9 of~\citet{caselli-etal-2021-guiding} in section \ref{casellietal}, we need the relationship typology to better reflect the notion of dynamics. To do so we propose an extension of the typology with another dimension -- \emph{growth}. To cover the differences in the involvement of these actor subcategories and the impact of a project on their development (as professionals or in society) we fragmented level 4 into three categories. These new levels assume that the hearing, HoH, and deaf researchers and the SL-users have equal positions, work on equal basis, with a shared amount of participation over the complete depth, breadth and scope into the life cycle of research and can maintain mutually beneficial collaborative network beyond the scope of a project. 

Another aspect that needs a more nuanced representation -- related to the depth and breadth of a research project -- relates to the direct and indirect impact of the project. NLP and MT projects address a very socially relevant and practically applied problem -- the problem of language. Following our preliminary literature analysis we noticed that the articles that do not involve the community have either a direct impact or use-case or have indirect impact or no immediate use-case. In general, any research output impacts with varying extent the user-community, which is even more so for the SLCs, as, for example, video recordings of signers lack privacy, or the use of 3D avatars links to ethical issues. Work that does not involve the users can either have a very big direct impact, e.g. being part of a bigger research,  have indirect and / or small impact. The development of sign language translation gloves which have a specific use-case without the deaf community is an example of the former case, while the development of a new method for sign language recognition for the purposes of MT is an example of the latter case. These variations should be acknowledged in a realistic, practically applicable typology as failing to do so would create a misguided classification or organization. To do so, in our typology we propose splitting level -1 in two: level -1.a covering the lack of or unwillingness to involvement of a diverse, representative user group when the product has a direct impact; and level -1.b covering the lack of involvement of a diverse, representative user group when the product has no direct impact or is part of a bigger project. As per~\citet{Harderetal2013}, -1.a may have a detrimental impact on society, while -1.b does not.


\begin{table}[]
    \hspace{-1cm}
    {\setlength{\tabcolsep}{2pt}
    \small
    \centering
    \begin{tabular}{|Y{1.5cm}|Y{1.5cm}|Y{1.5cm}|Y{1.5cm}|Y{1.5cm}|Y{1.5cm}|Y{1.5cm}|Y{1.5cm}|Y{1.5cm}|}\hline
         \multicolumn{2}{|c|}{Level (-1)} &  Level (0) & Level (1) & Level (2) & Level (3) & \multicolumn{3}{c|}{Level (4)} \\\hline
         \textit{Denigration direct impact} & \textit{Denigration indirect impact} & \textit{Neglect} & \textit{Learning about} & \textit{Learning from} & \textit{Learning together} & \textit{Learning as one} & \textit{Growing as one} & \textit{Working as one} \\\hline
         Hearing researchers make decisions without the SLC (neither HoH or deaf researchers) involvement, contrary to the SLCs interests, producing outputs with direct impact on the SLC. & Hearing researchers make decisions without the SLC (neither HoH or deaf researchers) involvement, contrary to or unaware of the SLCs interests, producing outputs with no direct impact on the SLC. & Hearing researchers make decisions without the SLC (neither HoH or Dear researchers) involvement, ignorant or dismissive of the SLCs interests. & Hearing researchers ask the SLCs and the users (and/or HoH or deaf researchers) opinions, but do not necessarily take them into account: the hearing researchers make the final decisions. & Hearing researchers ask the SLCs and the users opinions and consider the SLCs and users seriously. Hearing researchers still makes the final decision based on the information, HoH and deaf researchers are asked for evaluation, but not included in the process. & Major objectives and issues are discussed / negotiated jointly involving hearing, HoH and deaf researchers, and SL-users. Most decisions are made jointly, e.g. by consensus-building. & A consortium that includes hearing, HoH and deaf researchers, and SLC members, jointly built, discuss relevant issues by having knowledge exchange (e.g. seminars on different topics from all involved communities). & Hearing, HoH and deaf researchers, and SL-users work together on equal basis, are all integrated into the scope of the research cycle, but the SL-user is not involved in the execution of each step and / or the societal diversity is not representative. & Hearing, HoH and deaf researchers, and SL-users have a full consensus about the practices, the design is a continuous process and both the hearing researchers as well as the SL-users are equally integrated into the scope, depth and breadth of the research project.\\\hline
    \end{tabular}
    }
    \caption{Advanced typology of relationships of participation between deaf, hearing and HoH participants -- both as users and as researchers / developers.}
    \label{tbl:new_typology_advanced}
\end{table}

\begin{itemize}
\item   \textbf{Level 4.a. Learning as one.}
Establishing collaboration between researchers and SL-users from the beginning of the project can maximize the knowledge exchange (e.g. seminars on different topics from both communities). When driven by a common goal such collaboration has the potential to produce outputs that are beneficial and relevant to all stakeholders who can, in parallel, acquire cross-disciplinary knowledge and expertise. In terms of scope, all types of stakeholder should be involved in all stages of the decision-making process.
However, in terms of depth, this level does not distinguish how much \emph{power} each stakeholder is having in the process. Furthermore, it is unclear how diverse the group of the consortium is, i.e. its breadth, who the (SL) user is, and above all, it is unclear what the knowledge transfer flow is, that is, it has the pitfall where it can be to the (hearing) researchers, without the reciprocal transfer to the larger user community (or society in general).
\item    \textbf{Level 4.b. Growing as one}. This level suggests that in addition to learning as one, as in 4.a, potential avenues for creating value arise. These opportunities emerge as a result of changes - particular in the \textit{power} dimension - that present new possibilities for innovation, profit and improvement. Thus, hearing researchers and SL-users work together on equal basis, are both integrated into the scope of the research cycle, and are presented with opportunities to grow (professionally and societal); however, the SL-user is not involved in the execution of every relevant step and/or the societal diversity is not representative. 
\item   \textbf{Level 4.c. Working as one.} Researchers and SL-users have a full consensus about the practices, the design is a continuous and reciprocal process, and both the hearing as well as the SL-users are equally integrated into the scope, depth and breadth of the research project.
\end{itemize}






\subsection{Alignment with participatory design guiding principles} \label{casellietal}
As outlined in \ref{sec:co-creation}, \citet{caselli-etal-2021-guiding} calls for a better involvement of language users in NLP research. They advocate for change in perspective from  \textit{language as data} to \textit{language as people}.
To do so, \citet{caselli-etal-2021-guiding} propose 9 guidelines inspired by participatory design. These are presented in detail in Appendix~\ref{sec:app_guidelines} and briefly listed here: 

\begin{enumerate}
    \item Participatory design is about consensus and conflict.    
    \item Design is an inherently disordered and unfinished process.
    \item Communities are often not determined \textit{a priori}.
    \item Data and communities are not separate things.
    \item Community involvement is not scraping.
    \item Never stop designing.
    \item Text is a means rather than an end.
    \item The thin red line between consent and intrusion.
    \item The need to combine research goals, funding and societal political dynamics. 
\end{enumerate} 

We align these guidelines with our 4-level typology. We aim to show how our framework embeds them and can be used to judge the extent to which these are considered in the assessment of a project.

\begin{enumerate}
    \item \textbf{Consensus and conflict} are embedded in the communication between A and B throughout levels 1 to 4, where, in level (1) there is barely any consensus and conflicts remain unresolved while in level (4) consensus is achieved and conflicts are resolved. 
    \item To capture the concept of a continuous, reflexive and ongoing \textbf{design process} our typology assumes a frequency and volume of knowledge exchange and user community expansion. Levels -1 and 0 are on the one far end, where such exchange is inexistent, while level 4 assumes exchange of various types of knowledge that cover the plethora of expertise and expansion of the community with the growth of the project.
    \item As per the previous point, the complete set of user communities are \textbf{not determined} \textit{a priori}, but rather through the development process.\footnote{An interesting example is the involvement of deaf-blind participants in the SignON project, which was not defined at the start of the project.}
    \item The assumption that \textbf{communities are only data providers} raises the question where the separation line between SL-user and researchers is, or in which cases the SL-user indeed only provides data. In the last case we can categorize this on level 2. Levels 3 and 4 imply that the community can be involved in data production but users can take other roles too. With levels -1 to 1 the community is not involved in data generation. 
    \item According to the \textbf{community involvement is not scraping} principle ethical, equal, respectful and reciprocal social interactions are necessary for the creation or development of a tool for a specific community. Ethical engagement and expectations management should be a process conducted on level 3 (as learning from each others needs) and level 4 (in discussion with each other).
    We further split level 4 into three categories. Our levels 4.b. and level 4.c assume working together as equals, with clear ethical practices already described; level 4.a. assumes these are still to be developed and set in place. Ideally, working on equal levels is the most desired arrangement, however in most of the current SLMT projects this step is not implemented nor discussed (evident from our analysis presented in Section~\ref{sec:let_review}).  
    \item As acknowledged above, the interaction with the community should be continuous and frequent in order to \textbf{never stop designing} for a better solution. By including SLCs technical and resource issues can be decreased and participants effort can be recognized as labor. 
    \item The original formulation of the 7$^{th}$ principle of \cite{caselli-etal-2021-guiding} \textit{\textbf{Text} is a means rather than an end}. In order to capture different modalities of language, e.g. text, audio, video, we rephrase this principle as \textbf{Language is a means rather than an end.} This principle can be reflected in levels 2 to 4. Within levels 1 and below the lack of communication and developing solutions without the involvement of the SLC utilise language data without reflecting on its impact on the community. This principle is most prominent in 
    level 4b (Growing as one) and level 4c (Working as one). We ought to note that in most of the current SLMT work, this principle is comparable with level 2 as the researchers need the SLC for a switch in perspective, or level 3, wherein both parties have a discussion and consensus about which perspective is followed.
    \item \textbf{The thin red line between consent and intrusion} is a principle embedded in the lower levels already -- in -1 and 0 --. This line is crossed as the development of technology without the proper involvement could be considered intrusion (plenty are the examples of intrusive technology such as SL gloves which is not accepted by SLCs); from level 1, as soon as some form of recognition of language as people is formed, and onward, this principle is being considered in its positive form. 
    \item The complex dynamics of funding (for projects that support co-creation with the community) as well as goal formation for the research projects, and the community itself impact collaboration.  Until recently, the majority of SLMT-projects are not supported by a national or international grands, and thus are localised within a research team. As such they fall on level 1 or level 2. For active and effective collaboration, e.g. level 4, this principle should transcend beyond our typology and be adopted by funding bodies and agencies. In our typology, we do not specifically integrate this principle. However, we acknowledge the need of forming collaborative teams for which a common framework with sufficient financial, societal and political support as a prerequisite for level 3 and 4 collaborations.
\end{enumerate}


\section{Literature review}\label{sec:let_review}
In 2023 and 2024, two review articles of SLMT were published~\cite{Núñez-Marcosetal2023} and~\cite{decoster2024}. These presented an overview of a large volume of literature on SLMT focusing on the technological solutions, different approaches and historical advancements. These two review articles provide an extensive technical overview of SLMT, but little is noted on the inclusion of the deaf community (and in general, the user community). To the best of our knowledge, these are the most complete overviews of related work to date. That is why, we conducted an additional overview and analyzed the SLMT-related papers that were considered in the articles by \cite{decoster2024} and \cite{Núñez-Marcosetal2023} from the perspective of SL-users involvement. We then categorized them according to our newly-proposed typology. 

\subsection{Selection and filtering criteria}
In total we reviewed 127 papers from \cite{decoster2024} and \cite{Núñez-Marcosetal2023} -- 57 and 70, accordingly.  

The initial analysis yielded a combined number of \textbf{193 articles}. Next, we screened these papers by reading the (sub)titles, abstract, and participants we filtered out 66 papers. The remaining \textbf{127} articles were thoroughly read to identify to what extent the SLC has been involved in the research life cycle. From our final analysis we excluded 16 more articles, resulting in 111 articles. Our criteria were:

\begin{itemize}
\item	The paper needs to be mentioned in \cite{decoster2024} or \cite{Núñez-Marcosetal2023};
\item	The paper needs to have open access; 
\item	The study should focus on SLMT;
\item   The study should focus on SLs, or on the translation from SLs to SpLs -or in reverse-, but not only on spoken languages;
\item   It must be clear in how much and to what extent the SL-user was involved.
\end{itemize} 

After these exclusion steps, the remaining \textbf{111 papers}, 57,5 \% of the original 193 papers were considered in the following discussion. We list the 111 papers in Appendix~\ref{sec:app_papers_table}.

In our analysis, we looked to what extent SL-users were involved into the different stages of the research and development described in these articles and how this involvement spreads-out over the newly-proposed typology. Our review unveiled that SL-users have not been involved beyond Level 2. Table~\ref{tbl:amountarticlesrev} shows the total amount of reviewed papers categorized over the different levels; in Section~\ref{sec:co-creation in reviewed articles} we present more detailed analysis of the literature review.

\begin{table}[!h]
    \begin{tabular}{|l|l|}\hline
   \textbf{Typological levels} & \textbf{Total of articles per level} \\\hline
    Level -1           & 83                          \\\hline
    Level 0            & 13                          \\\hline
    Level 1            & 13                          \\\hline
    Level 2            & 2  \\\hline  
    Level 3            & 0  \\\hline 
    Level 4            & 0  \\\hline\hline
    Total              & 111\\\hline
    \end{tabular}
\caption{The amount of reviewed articles per level}
    \label{tbl:amountarticlesrev}
\end{table}

\subsection{Distribution of articles over levels of involvement} \label{sec:co-creation in reviewed articles}
We analyzed the SL-users involvement in the different stages of the projects described in \textbf{111 papers}. We ought to note that despite the negative connotation of levels -1 and 0, our typology merely assesses the process of co-creation and does not question the validity or the robustness of the conducted research, except if explicitly noted so. Table~\ref{tbl:amountarticlesrev} gives an overview of the total amount of reviewed papers categorized over the different participation levels. In general, what we observe is that the SL-users have not been involved into Level 3 and Level 4.

Below, we outline how the articles were categorized according to the typological levels, provide examples of common assumptions made to explain this categorization, and discuss how the SLC and SL-users could have been engaged in a more ethical and efficient manner. 


Articles presenting work that \emph{only} focuses on the MT process or on the model and technique's comparison without the involvement of SL-users (nor assessing the impact on the user community in general) and is conducted only by hearing researchers, were categorized at level -1. As described in Section~\ref{sec:co-creation-sLMT}, this level includes hearing researchers who make decisions without the SLC being involved (and sometimes contrary to the SLCs interests). In 75\% (i.e. 83 out of 111 articles) of the reviewed articles, SL-users were not involved in any of the tasks or stages of the research life cycle. An example of this is the study of \cite{Fangetal2017}, in which 11 hearing participants were asked to learn ASL signs in a 3-hour tutorial; these were then recorded and used in the development of the proposed method and its analysis. The implementation of the proposed method and its analysis not only rely on data that is not representative of natural (American) SL, but also overlook the complexity of SLs and the needs of SL communities. The assumption that watching a 3-hour tutorial is sufficient to learn a new language in a different modality\footnote{Despite the limited scope of the presented work, SLs have extensive vocabulary, complex grammars, and even more, operate on different simultaneous produced parameters, e.g. hands, face, torso, etc.} brings into question the validity and robustness of the research results. Additionally, it raises ethical concerns, suggesting the potential marginalization and discrimination of the SLC. This example is therefore categorized as Level -1a. As discussed in Section \ref{advanced typology}, work that does not involve the users, can have a direct impact or an indirect impact on the user community (and in general on society): level -1a, or level 1b. From the 83 articles categorized under level -1, 4 are labeled as -1a and 79 are as -1b (see Table~\ref{tbl:amountarticleslevel-1ab}). 

\begin{table}[!h] 
    \begin{tabular}{|l|l|}\hline
   \textbf{Typological level} & 
   \textbf{Number of articles} \\\hline
    Level -1a           &4                          \\\hline
    Level -1b           &79                          \\\hline
    Total              &83\\\hline
    \end{tabular}
\caption{The categorization of level -1 over the reviewed articles}
    \label{tbl:amountarticleslevel-1ab}
\end{table} 

In less than 4.8 \% (e.g. 4 of the 83 articles) of the same cases, there are human participants (that are not the researchers themselves) involved who are not representative for the primary user group, i.e. the SLCs. An example of this is \cite{Forster2012}, wherein the authors created the RWTH-Phoenix-weather database, based on hearing interpreter services. We categorize this under Level -1a as (1) the data is not containing natural sign language production, as it is highly influenced by written text, (2) the outputs of the data are used in further formalizations and systems of different MT-topics related to recognition and/or production which makes further research based on this specific database less reliable, and (3) the recordings should have been provided by deaf interpreters (instead of hearing interpreters) as the translation was mostly based on written text, and 4) the impact on the SLC is direct.

\ 

At level 0 we consider research that involves SL-users, but their participation is limited to tasks such as data collection, recording, or annotation, without contributing to ideation, research, or development. SL-users are not involved in the decision-making; the research team is comprised only of non-signers. Based on this criterion, we conclude that 13 out of the 111 reviewed articles, i.e. 11.7\%, are at level 0. An example is the article of by \cite{Morrissey2011}, who included one native Irish SL-signer for data collection -- for the recordings of the dialogue and its manual translation. While the concept of recording a native signer to create a corpus is valuable, as it captures natural language production, the number of participants raises concerns. With only one participant, it is difficult to capture language variation. Not only dialectal variants, but also variations related to diversity and educational level of the participants.
Another example is the work of \citet{MassoeBadia2010}. While they create a corpus involving a native deaf signer, their corpus creation approach is problematic. First, it is a written text (that is, originally created in spoken language) which is then translated into sign language, thus the signed content is not original which according to \cite{desisto-EtAl:2022:LREC} is a suboptimal situation. Second, they involved only one person, and therefore did not take into account sign variations. And third, there is no information about whether this person was familiar with the domain of translation which can impact the translation output. Additionally, their experiments did not include human evaluation, which further undermines the validity of their results. For this reason, we categorize their paper at level 0.

The criteria for level 1 are to contain deaf signers involvement in both content / data creation, e.g. recordings or annotations, as well as in system assessment and feedback on the system (without their direct involvement in the development process). There are 13 articles, i.e. 11.7\% which fall in this category. In these articles, the authors have different approaches: some contacted early deaf students \cite{Chiuetal2006} to grade the utility of the proposed approach, while others made a combination in between hearing evaluators that are experienced with a signed language and native signers \cite{Wuetal2007} for the evaluation of the translated sentences. In the study of \cite{su_wu2009}, the authors developed a bilingual corpus annotated and verified by SL-linguists, and involved 10 deaf students in the evaluation process. What these example show, is that the SLC is involved in one or more different tasks over the \textit{breadth} and \textit{scope} in the research cycle. 

Level 2 builds on Level 1 with the addition of the criterion that different \textit{National Associations of the Deaf} (NADs) are involved. These organizations are specialized in their SLCs, and therefore can provide better expertise, guidance and communication with, for, and about the SLC. Only 2 out of the 111 papers align with this criterion, i.e. less then 1.8\%. In both of the articles of \cite{Morrissey2007} and \cite{Jantunenetal2021} NADs has been involved. The approaches differ: in \cite{Morrissey2007}, deaf colleagues and members of the SLCs were contacted via Deaf Studies university programs to make a choice of a domain for SLT, asking their cooperation for the human translation, advice on the SL grammar and linguistics, and to evaluate the translated output. \citet{Jantunenetal2021} also devoted a section to co-engineering, participation and culture. 

For an article to be classified at Level 3 or at Level 4 the research should have involved the SLCs or SL-users in the decision-making process for an agreement based on discussion (level 3) and have a balanced research team or consortium composed of signers and non-signers (level 4). None of the articles we reviewed fulfilled these criteria.

\section{Proposal of formal guidelines for co-creation in SLMT projects}\label{sec:formal guidelines}
Following the review of co-creation applied in different fields (including NLP) summarized in Section~\ref{sec:co-creation}, the newly-proposed typology (Section~\ref{sec:co-creation-sLMT}) and the review of SLMT-related literature in Section~\ref{sec:let_review}, here we set-up a co-creation guidelines framework for SLP-projects to reach Level (4a), Level (4b), or even Level (4c). 



\subsection{Challenges}
\begin{enumerate}
    \item \textbf{Positionality and privileges of hearing, non-signing \footnote{There are other forms of biases that should be avoided. However, these are not in the scope of our work and therefore not discussed here.} researchers}. As we noted in our literature review (Section~\ref{sec:let_review}), sign language projects have been led by hearing, non-signing individuals creating bias in the landscape of SLMT research. Although established traditions in the field are still favoring the aforementioned group of researchers, we are noticing a shift towards more inclusive research, which should be promoted and needs to become the default practice. As in the study of~\cite{Holcomb2024}, our study shows that while some individuals showed curiosity to deep knowledge and understanding of existing biased and systematic oppression, the \textbf{111} reviewed papers involved technical researchers who were holding up existing power structures. As a consequence, they were excluding and hindering the SLC, and SL-user from their research process.  
    \item \textbf{Inclusion for the sake of inclusion}. Including SL-users without taking seriously their unique perspectives in a mostly hearing-led projects is an ineffective and unethical practice, which we place at level 0. As~\citet{Holcomb2024} note:
    
    \begin{quote}
     deafness is earnestly viewed as a benefit and a valuable contribution to the world, a concept known as ``deaf Gain''. In other words, it is argued that comparing hearing people to deaf people should be understood as comparing apples to oranges. Each has its own unique advantages and disadvantages, but both are valuable, can thrive in environments that support their natures, and can enrich the human experience in positive ways.~\cite{Holcomb2024}
    \end{quote}
    
    \item \textbf{Size of the user population.} SLC, and the SL-user in general, are a small sub-population. As such, many individuals been requested for such kind of projects over and over again. This leads to \textit{research fatigue} where the same population is requested to participate in technical projects, without getting the benefit of their contributions, the SL-user becomes tired for these kind of requests \cite{Meulder2024}.

    \item \textbf{Adhering to ethical protocols.} While in our work we addressed the topic of co-creation (with SL-users and the SLC), we did not yet mention the \textit{ethical protocols} for research in or with the SLC. Such need to be placed for any research involving human participants and therefore, SLMT projects require the assessment of ethical committees prior to their commencement so that ethical, fair, transparent and sustainable collaboration with linguistic and cultural minority groups is ensured. 
\end{enumerate}

\subsection{Proposals} 
\begin{enumerate}
    \item We support \citet{Harderetal2013}'s suggestion and encourage researchers to tailor our newly proposed typology to their specific use case by defining the initial participants in groups A and B, while remaining flexible in adjusting these groups as the project progresses.
    \item We propose that ongoing and planned activities include regular self-evaluations based on the proposed typology to assess the level at which their work is categorized.
    \item Conduct research in cross-disciplinary and multi-disciplinary SLP fields. We recommend to start the conversation with the SL-users and SLC early in the project, e.g. in the ideation phase, as well building up a network with other (deaf) researchers or disciplines, such as for example deaf studies and NADs. 
    \item As mentioned in Section~\ref{sec:co-creation-sLMT}, the original typology of \cite{Harderetal2013} does not allow us to show the proportions or \textit{power} between hearing, and HoH or deaf researchers, as the focus is mainly on co-creation with the society. Our typology is based on the history of technological hearing-led projects, and we categorized deaf and Hard-of-Hearing researchers under the concept of SLCs. This is obviously incorrect, as showed in our typology in Table~\ref{tbl:new_typology_advanced} by the distinction of Level 2 to Level 3, and from Level 3 to Level 4, wherein we slightly shift from `tasks' in-between the levels and the implementation of HoH and deaf researchers, and the addition of levels 4b and 4c (along with 4a): Learning as one, which still implicates the categorization of HoH/deaf versus hearing researchers and the SL-users, while the Level `Growing as one' and further on has put these three categories (hearing, HoH or deaf researchers) in one box with as an opposite contrast the SLC. We suggest creating consortia with fair involvement of these type of researchers on strategic positions, moving beyond the dominantly-hearing consortia. However, the included participants should have an active role, avoiding the challenge of ``inclusion for the sake of inclusion''.
    \item Discuss and establish common goals to allow all participants (researchers and users) including hearing, HoH and deaf individuals on equal basis to benefit.
    \item Establish communication and dissemination protocols from the beginning. This would involve hiring interpreters and therefore budget should be provisioned from the inception of a project. Furthermore, communication and dissemination should be conducted in a language which the participants are fluent with in order to reduce miscommunication and misunderstandings. Timeline communication can also manage expectations leading to achievable goals.
    \item Expand the breadth, depth and scope of the project during its evolution through including more users and advancing the technology to address their needs.
    \item Secure ethical approvals from research ethics committees. Value the privacy of participants and respect their wishes (e.g. in the case of participation withdrawal). Be aware of research fatigue. 
    \item Co-creation is a dynamic process and changes should be welcome. 
    \item Continuous assessment is beneficial for expectation management and alignment of participants and goals. 
\end{enumerate}



\section{Conclusions}
This study investigates the topic of co-creation for SLMT research. After reviewing existing work on co-creation as defined in different fields, and using the work of~\citet{caselli-etal-2021-guiding} and~\citet{Harderetal2013} we developed a participation typology for the involvement of SLCs in SLNLP and SLMT research. We then applied this typology to assess 111 existing articles on the topic of SLMT. Our study showed that the articles we reviewed, were (mainly) centered around a technical perspective, and that language has been seen as data, in contrast to the proposal of \cite{caselli-etal-2021-guiding} to see language as people, rather than data; most of the \textbf{111} reviewed articles, were placed on Level -1 (Denigration), Level 0 (Neglect) or Level 1 (Learning From). 
That is, none of the reviewed articles involved SL-users and deaf researchers at Level 3 or Level 4 of our proposed typology. Furthermore, the number of papers decreases with the increase of the level number (i.e. Level -1, 83 articles, Level 0, 13 articles, Level 1, 13 articles, Level 2, 2 articles). This is an indication that a change has been set in motion, in which collaboration with the SLC is increasing, but that the tools for a fair, ethical and responsible collaboration and co-creation are missing.  

Guided by the advanced typology, we proposed a set of 9 guidelines to improve the SLMT research landscape (from a co-creative perspective); we noted 4 overarching challenges that these guidelines tackle. These principles should be followed in parallel with the ones by~\citet{caselli-etal-2021-guiding}. Finally, it is worth stressing the dynamic nature of co-creation. As projects evolve, so do communities and their requirements. We did not explore this temporal aspect in detail and have left it for future work, recognizing the importance of considering changing requirements.

\vspace{6pt}

\abbreviations{Abbreviations}{
The following abbreviations are used in this paper:\\

\noindent 
\begin{tabular}{@{}ll}
CBPR & Community-Based Participatory Research\\
CPUs & Central Processing Unit\\
GPUs & Graphics Processing Unit\\
HoH & Hard of Hearing\\
LLMs & Large Language Models\\
MT & Machine Translation\\
NADs & National Associations for the deaf\\
NLP & Natural Language Processing\\
PD & Participatory Design\\
SLs & Sign Languages\\
SLCs & Sign Language Communities\\
SLMT & Sign Language Machine Translation\\
SLNLP & Sign Language Natural Language Processing\\
SLP & Sign Language Processing\\
SpLs & Spoken Languages\\
\end{tabular}
}

\appendixtitles{no} 
\appendixstart
\appendix

\section[\appendixname~\thesection]{The nine guidelines of~\citet{caselli-etal-2021-guiding}}\label{sec:app_guidelines}

\begin{enumerate}
    \item \textbf{PD is about consensus and conflict.} The design of co-creation should be conducted in discussion and alignment between the involved parties. 
    \item \textbf{Design is an inherently disordered and unfinished process.} The design should be a continuous, reflexive and ongoing process (principle 2 and 6 of \cite{caselli-etal-2021-guiding} and level 4c of our proposed typology in Table~\ref{tbl:new_typology_advanced}.  \cite{caselli-etal-2021-guiding} mention that the term \textit{community} needs to be defined in a reflexive and adaptable manner, with its continuous changes.\cite{Harderetal2013} assume that this definition is a fixed format, based on the amount of \textit{power} of different researchers (i.e. hearing, HoH or deaf) to define the SLC.     
    \item Communities are often \textbf{not determined} \textit{a priori}.
    \item \textbf{Data and communities are not separate things} Principle 4 of \cite{caselli-etal-2021-guiding} contain the assumption that we expect that communities have a prominent role in the development of NLP-systems, but that the communities until now most often only function as language data providers. This assumption raises the question where the separation line between SL-user and researchers is, or in which cases the SL-user indeed only provides data. In the last case we can categorize this on level 2 of \cite{Harderetal2013}.
    \item \textbf{Community involvement is not scraping} In principle 5, the social interactions are described as necessary for the  creation or development of a tool for a specific community, wherein also the ethical engagements, equity, reciprocity, and respect should be discussed. As Level 4.b. and level 4.c assume that working together in equality, with clear ethical practices are already described, this principle is also hard to divide to one level. Ideally suited -yes- working on equal level is the highest possible achievement, although in most of the current SLMT projects this step is not implemented or discussed. The development of the expectations/ ethical engagement should be on level 3 (as this part is meant as learning from each others needs) or level 4 (in discussion with each other), and if this is already discussed and decided, then this principle can be divided into level 4b or level 4c for the execution. But also in this case, a reciprocity attitude is needed for reflection and adaption of execution.  
    \item \textbf{Never stop designing} Principle 6 states out that when a NLP-tool is based on PD, there should be awareness about the needs of the SLC and include them into the design stage. By including them, technical and resource issues can be decreased, and participants effort can be recognized as labor. 
    \item \textbf{Language\footnote{Please be aware that in the article of \cite{caselli-etal-2021-guiding} the original principle is \textit{Text is a means rather than an end}, that we have more specified in this article to language.} is a means rather than an end.} Principle 7 refers to switch the perspective from \textit{language as data} to \textit{language as people}, wherein the main focus should be to serve people's needs instead of trying to copy people's language use. This principle can ideally be compared with level 4b (Growing as one) or level 4c (Working as one), but in most of the current SLMT this principle is comparable with level 2 -as the researchers need the SLC for this perspective-switch- or level 3, wherein both parties have a discussion and consensus about which perspective is followed.
    \item \textbf{The thin red line between consent and intrusion} Principle 8 can be part of some of the lower levels already - as soon as some form of recognition of language as people is formed, so this principle can be seen as 'learning about' (level 1) or 'Learning from (Level 2). 
    \item \textbf{The need to combine research goals, funding and societal political dynamics.} The last principle - principle 9 - refers to the complex dynamics of funding (for projects that support co-creation with the community), goals of the research projects, and the community itself. As the most SLMT-projects are not supported by a grand for the above needed adaptations, this principle can be compared to level 1 or level 2. 
\end{enumerate}

\section[\appendixname~\thesection]{Paper reviews}\label{sec:app_papers_table}

\begin{enumerate}
{\small 
    \item	Angelova, G., Avramidis, E., Möller, S.: Using neural machine translation methods for sign language translation. In: Proceedings of the 60th Annual Meeting of the Association for Computational Linguistics: Student Research Workshop, pp. 273–284 (2022)	Deaf involvement: 	no	Evaluation: there are no deaf people involved: only focus on the MT-process. Level -1	Level: 	-1	b	Depth: 	Only (hearing) researchers	Breadth: 	No different groups/ variations are involved	Scope: 	In none of the research life cycle steps
\item	Arvanitis, N., Constantinopoulos, C., Kosmopoulos, D.: Translation of sign language glosses to text using sequence-to-sequence  attention models. In: 2019 15th International Conference on  Signal-Image Technology \& Internet-Based Systems (SITIS),  pp. 296–302 (2019). \url{https://doi.org/10.1109/SITIS.2019.00056}.  IEEE	Deaf involvement: 	no	Evaluation: there are no deaf people involved: only focus on the MT-process. Level -1	Level: 	-1	b	Depth: 	Only (hearing) researchers	Breadth: 	No different groups/ variations are involved	Scope: 	In none of the research life cycle steps
\item	Barberis, D., Garazzino, N., Prinetto, P., Tiotto, G., Savino, A., Shoaib, U., et al. (2011). Language resources for computer assisted translation from italian to italian sign language of deaf people. In Proceedings of accessibility reaching everywhere AEGIS workshop and international conference (pp. 96–104).	Deaf involvement: 	Yes, an interpret that helped in the production of signs	Evaluation: there are no deaf people involved: only focus on the MT-process. Level -1	Level: 	-1	b	Depth: 	Only (hearing) researchers	Breadth: 	No different groups/ variations are involved	Scope: 	In none of the research life cycle steps
\item	Bauer, B., Nießen, S., \& Hienz, H. (1999). Towards an automatic sign language translation system. In In 1st international. Citeseer.	Deaf involvement: 	Yes, 1 DGS interpreter	Evaluation: there are no deaf people involved, only one hearing interpreter for data recordings/ collection	Level: 	-1	b	Depth: 	Only (hearing) researchers	Breadth: 	No different groups/ variations are involved	Scope: 	In none of the research life cycle steps
\item	Brour, M., \& Benabbou, A. (2019). ATLASLang MTS 1: Arabic text language into Arabic Sign Language machine translation system. Procedia Computer Science, 148, 236–245.	Deaf involvement: 	No	Evaluation: there are no deaf people involved: only focus on the MT-process. Level -1	Level: 	-1	b	Depth: 	Only (hearing) researchers	Breadth: 	No different groups/ variations are involved	Scope: 	In none of the research life cycle steps
\item	Bungeroth, J., Ney, H.: Statistical sign language translation.  In: Workshop on Representation and Processing of Sign Languages, LREC, vol. 4, pp. 105–108 (2004). Citese	Deaf involvement: 	No	Evaluation: there are no deaf people involved: only focus on the MT-process. Level -1	Level: 	-1	b	Depth: 	Only (hearing) researchers	Breadth: 	No different groups/ variations are involved	Scope: 	In none of the research life cycle steps
\item	Camgoz, N. C., Koller, O., Hadfield, S., \& Bowden, R. (2020a). Multi-channel trans- formers for multi-articulatory sign language translation. In European conference on computer vision (pp. 301–319). Springer.	Deaf involvement: 	No	Evaluation: there are no deaf people involved: only focus on the MT-process. Level -1	Level: 	-1	b	Depth: 	Only (hearing) researchers	Breadth: 	No different groups/ variations are involved	Scope: 	In none of the research life cycle steps
\item	Camgoz, N.C., Koller, O., Hadfield, S., Bowden, R.: Sign lan- guage transformers: Joint end-to-end sign language recognition and translation. In: Proceedings of the IEEE/CVF Conference on Computer Vision and Pattern Recognition, pp. 10023–10033 (2020)	Deaf involvement: 	Yes, the existing 9 DGS-signers of the dataset	Evaluation: there are no deaf people involved: only focus on the MT-process. Level -1	Level: 	-1	b	Depth: 	Only (hearing) researchers	Breadth: 	No different groups/ variations are involved	Scope: 	In none of the research life cycle steps
\item	Cao, Y., Li, W., Li, X., Chen, M., Chen, G., Hu, L., et al. (2022). Explore more guidance: A task-aware instruction network for sign language translation enhanced with data augmentation. arXiv preprint arXiv:2204.05953.	Deaf involvement: 	no	Evaluation: there are no deaf people involved: only focus on the MT-process. Level -1	Level: 	-1	b	Depth: 	Only (hearing) researchers	Breadth: 	No different groups/ variations are involved	Scope: 	In none of the research life cycle steps
\item	Chaudhary, L., Ananthanarayana, T., Hoq, E., Nwogu, I.: Signnet ii: A transformer-based two-way sign language translation model. IEEE Transactions on Pattern Analysis and Machine Intelligence (2022)	Deaf involvement: 	no	Evaluation: there are no deaf people involved: only focus on the MT-process. Level -1	Level: 	-1	b	Depth: 	Only (hearing) researchers	Breadth: 	No different groups/ variations are involved	Scope: 	In none of the research life cycle steps
\item	Chen, Y., Wei, F., Sun, X., Wu, Z., \& Lin, S. (2022). A simple multi-modality transfer learning baseline for sign language translation. In Proceedings of the IEEE/CVF conference on computer vision and pattern recognition (pp. 5120–5130).	Deaf involvement: 	no	Evaluation: there are no deaf people involved: only focus on the MT-process. Level -1	Level: 	-1	b	Depth: 	Only (hearing) researchers	Breadth: 	No different groups/ variations are involved	Scope: 	In none of the research life cycle steps
\item	Chen, Y., Zuo, R., Wei, F., Wu, Y., Liu, S., Mak, B.: Two-stream network for sign language recognition and translation. arXiv pre- print arXiv: 2211. 01367 (2022)	Deaf involvement: 	no	Evaluation: there are no deaf people involved: only focus on the MT-process. Level -1	Level: 	-1	b	Depth: 	Only (hearing) researchers	Breadth: 	No different groups/ variations are involved	Scope: 	In none of the research life cycle steps
\item	D’Haro, L. F., San-Segundo, R., Cordoba, R. d., Bungeroth, J., Stein, D., \& Ney, H. (2008). Language model adaptation for a speech to sign language translation system using web frequencies and a map framework. In Ninth annual conference of the international speech communication association.	Deaf involvement: 	no	Evaluation: there are no deaf people involved: only focus on the MT-process. Level -1	Level: 	-1	b	Depth: 	Only (hearing) researchers	Breadth: 	No different groups/ variations are involved	Scope: 	In none of the research life cycle steps
\item	Dasgupta, T., \& Basu, A. (2008). Prototype machine translation system from text-to- Indian sign language. In Proceedings of the 13th international conference on intelligent user interfaces (pp. 313–316).	Deaf involvement: 	No	Evaluation: we have evaluated the sys?tem based on the feedbacks of the ISL experts	Level: 	-1	b	Depth: 	Only (hearing) researchers	Breadth: 	No different groups/ variations are involved	Scope: 	In none of the research life cycle steps
\item	Davydov, M., \& Lozynska, O. (2017a). Information system for translation into Ukrainian sign language on mobile devices. In 2017 12th international scientific and technical conference on computer sciences and information technologies, Vol. 1 CSIT, (pp. 48–51). IEEE.	Deaf involvement: 	No	Evaluation: there are no deaf people involved: only focus on the MT-process. Level -1	Level: 	-1	b	Depth: 	Only (hearing) researchers	Breadth: 	No different groups/ variations are involved	Scope: 	In none of the research life cycle steps
\item	De Coster, M., D’Oosterlinck, K., Pizurica, M., Rabaey, P., Ver- linden, S., Van Herreweghe, M., Dambre, J.: Frozen pretrained transformers for neural sign language translation. In: Proceedings of the 1st International Workshop on Automatic Translation for Signed and Spoken Languages (AT4SSL), pp. 88–97. Associa- tion for Machine Translation in the Americas, Virtual (2021).	Deaf involvement: 	No	Evaluation: there are no deaf people involved: only focus on the MT-process. Level -1, but the authors mention that 'co-creation with the DHH community members is the key'. 	Level: 	-1	b	Depth: 	Only (hearing) researchers	Breadth: 	No different groups/ variations are involved	Scope: 	In none of the research life cycle steps
\item	De Coster, M., Dambre, J.: Leveraging frozen pretrained written language models for neural sign language translation. Informa- tion 13(5), 220 (2022)	Deaf involvement: 	no	Evaluation: there are no deaf people involved: only focus on the MT-process. Level -1	Level: 	-1	b	Depth: 	Only (hearing) researchers	Breadth: 	No different groups/ variations are involved	Scope: 	In none of the research life cycle steps
\item	Dey, S., Pal, A., Chaabani, C., Koller, O.: Clean text and full- body transformer: Microsoft’s submission to the wmt22 shared task on sign language translation. In: Proceedings of the Seventh Conference on Machine Translation, pp. 969–976. Association for Computational Linguistics, Abu Dhabi (2022). \url{https://aclanthology.org/2022.wmt-1.93}	Deaf involvement: 	No (at least not clear mentioned: the authors mention something about human evaluation, but it seems that that is out of the scope of this article).	Evaluation: there are no deaf people involved: only focus on the MT-process. Level -1	Level: 	-1	b	Depth: 	Only (hearing) researchers	Breadth: 	No different groups/ variations are involved	Scope: 	In none of the research life cycle steps
\item	Dreuw, P., Stein, D., Deselaers, T., Rybach, D., Zahedi, M., Bungeroth, J., Ney, H.: Spoken language processing techniques for sign language recognition and translation. Technol. Disabil. 20(2), 121–133 (2008)	Deaf involvement: 	no	Evaluation: there are no deaf people involved: only focus on the MT-process. Level -1	Level: 	-1	b	Depth: 	Only (hearing) researchers	Breadth: 	No different groups/ variations are involved	Scope: 	In none of the research life cycle steps
\item	Dreuw, P., Stein, D., Deselaers, T., Rybach, D., Zahedi, M., Bungeroth, J., et al. (2008). Spoken language processing techniques for sign language recognition and translation. Technology and Disability, 20(2), 121–133.	Deaf involvement: 	no	Evaluation: there are no deaf people involved: only focus on the MT-process. Level -1	Level: 	-1	b	Depth: 	Only (hearing) researchers	Breadth: 	No different groups/ variations are involved	Scope: 	In none of the research life cycle steps
\item	Dreuw, P., Stein, D., Ney, H.: Enhancing a sign language transla- tion system with vision-based features. In: International Gesture Workshop, pp. 108–113 (2007). Springer	Deaf involvement: 	no	Evaluation: there are no deaf people involved: only focus on the MT-process. Level -1	Level: 	-1	b	Depth: 	Only (hearing) researchers	Breadth: 	No different groups/ variations are involved	Scope: 	In none of the research life cycle steps
\item	Egea, S., McGill, E., \& Saggion, H. (2021). Syntax-aware transformers for neural machine translation: The case of text to sign gloss translation. In Proceedings of the 14th workshop on building and using comparable corpora.	Deaf involvement: 	no	Evaluation: there are no deaf people involved: only focus on the MT-process. Level -1	Level: 	-1	b	Depth: 	Only (hearing) researchers	Breadth: 	No different groups/ variations are involved	Scope: 	In none of the research life cycle steps
\item	Fang, B., Co, J., \& Zhang, M. (2017). Deepasl: Enabling ubiquitous and non-intrusive word and sentence-level sign language translation. In Proceedings of the 15th ACM conference on embedded network sensor systems (pp. 1–13).	Deaf involvement: 	11 hearing participants who learned ASL via 3-hours tutorials	Evaluation: level -1: contrary to the SLCs interests) 11 hearing participants who learnerd ASL via 3-hours tutorials	Level: 	-1	a	Depth: 	Only (hearing) researchers	Breadth: 	No different groups/ variations are involved	Scope: 	In none of the research life cycle steps
\item	Foong, O. M., Low, T. J., \& La, W. W. (2009). V2s: Voice to sign language translation system for malaysian deaf people. In International visual informatics conference (pp. 868–876). Springer.	Deaf involvement: 	Yes, 100 people (groups of children, male, female, young and older), but no deaf.	Evaluation: It is not focused on SL, but on SpLs	Level: 	-1	b	Depth: 	Only (hearing) researchers	Breadth: 	No different groups / variations are involved	Scope: 	In none of the research life cycle steps
\item	Forster, J., Schmidt, C., Hoyoux, T., Koller, O., Zelle, U.,  Piater, J.H., Ney, H.: Rwth-phoenix-weather: A large vocabu?lary sign language recognition and translation corpus. In:  LREC, vol. 9, pp. 3785–3789 (2012)	Deaf involvement: 	It was not implemented	Evaluation: there are no deaf people involved: only focus on the MT-process. Level -1	Level: 	-1	a	Depth: 	Only (hearing) researchers	Breadth: 	No different groups/ variations are involved	Scope: 	In none of the research life cycle steps
\item	Forster, J., Schmidt, C., Koller, O., Bellgardt, M., Ney, H.: Exten- sions of the sign language recognition and translation corpus rwth-phoenix-weather. In: LREC, pp. 1911–1916 (2014)	Deaf involvement: 	no	Evaluation: there are no deaf people involved: only focus on the MT-process. Level -1	Level: 	-1	b	Depth: 	Only (hearing) researchers	Breadth: 	No different groups/ variations are involved	Scope: 	In none of the research life cycle steps
\item	Fu, B., Ye, P., Zhang, L., Yu, P., Hu, C., Chen, Y., et al. (2022). ConSLT: A token- level contrastive framework for sign language translation. arXiv preprint arXiv: 2204.04916.	Deaf involvement: 	no	Evaluation: there are no deaf people involved: only focus on the MT-process. Level -1	Level: 	-1	b	Depth: 	Only (hearing) researchers	Breadth: 	No different groups/ variations are involved	Scope: 	In none of the research life cycle steps
\item	Gan, S., Yin, Y., Jiang, Z., Xie, L., Lu, S.: Skeleton-aware neu- ral sign language translation. In: Proceedings of the 29th ACM International Conference on Multimedia, pp. 4353–4361 (2021)	Deaf involvement: 	no	Evaluation: there are no deaf people involved: only focus on the MT-process. Level -1	Level: 	-1	b	Depth: 	Only (hearing) researchers	Breadth: 	No different groups/ variations are involved	Scope: 	In none of the research life cycle steps
\item	Grieve-Smith, A. B. (1999). English to American Sign Language machine translation of weather reports. In Proceedings of the second high desert student conference in linguistics (HDSL2), Albuquerque, NM (pp. 23–30).	Deaf involvement: 	No, althouh the author mention in future work that the ouput needs to be cross-checked with a native signer	Evaluation: there are no deaf people involved: only focus on the MT-process. Level -1	Level: 	-1	b	Depth: 	Only (hearing) researchers	Breadth: 	No different groups/ variations are involved	Scope: 	In none of the research life cycle steps
\item	Grif, M. G., Korolkova, O. O., Demyanenko, Y. A., \& Tsoy, Y. B. (2011). Development of computer sign language translation technology for deaf people. In Proceedings of 2011 6th international forum on strategic technology, Vol. 2 (pp. 674–677). IEEE.	Deaf involvement: 	not clear	Evaluation: there are no deaf people involved: only focus on the MT-process. Level -1	Level: 	-1	b	Depth: 	Only (hearing) researchers	Breadth: 	No different groups/ variations are involved	Scope: 	In none of the research life cycle steps
\item	Guo, D., Zhou, W., Li, A., Li, H., \& Wang, M. (2019). Hierarchical recurrent deep fusion using adaptive clip summarization for sign language translation. IEEE Transactions on Image Processing, 29, 1575–1590.	Deaf involvement: 		Evaluation: there are no deaf people involved: only focus on the MT-process. Level -1	Level: 	-1	b	Depth: 	Only (hearing) researchers	Breadth: 	No different groups/ variations are involved	Scope: 	In none of the research life cycle steps
\item	Halawani, S. M. (2008). Arabic sign language translation system on mobile de- vices. IJCSNS International Journal of Computer Science and Network Security, 8(1), 251–256.	Deaf involvement: 	No	Evaluation: there are no deaf people involved: only focus on the MT-process. Level -1	Level: 	-1	b	Depth: 	Only (hearing) researchers	Breadth: 	No different groups/ variations are involved	Scope: 	In none of the research life cycle steps
\item	Hoque, M. T., Rifat-Ut-Tauwab, M., Kabir, M. F., Sarker, F., Huda, M. N., \& Abdullah-Al- Mamun, K. (2016). Automated bangla sign language translation system: Prospects, limitations and applications. In 2016 5th international conference on informatics, electronics and vision ICIEV, (pp. 856–862). IEEE.	Deaf involvement: 	no	Evaluation: there are no deaf people involved: only focus on the MT-process. Level -1	Level: 	-1	a	Depth: 	Only (hearing) researchers	Breadth: 	No different groups/ variations are involved	Scope: 	In none of the research life cycle steps
\item	Huang, J., Zhou, W., Zhang, Q., Li, H., Li, W.: Video-based  sign language recognition without temporal segmentation. In:  Proceedings of the AAAI Conference on Artifcial Intelligence,  vol. 32 (2018)	Deaf involvement: 	no	Evaluation: there are no deaf people involved: only focus on the MT-process. Level -1	Level: 	-1	b	Depth: 	Only (hearing) researchers	Breadth: 	No different groups/ variations are involved	Scope: 	In none of the research life cycle steps
\item	Huenerfauth, M. (2004). A multi-path architecture for machine translation of english text into American Sign language animation. In Proceedings of the student research workshop at HLT-NAACL 2004 (pp. 25–30).	Deaf involvement: 	no	Evaluation: there are no deaf people involved: only focus on the MT-process. Level -1	Level: 	-1	b	Depth: 	Only (hearing) researchers	Breadth: 	No different groups/ variations are involved	Scope: 	In none of the research life cycle steps
\item	Jin, T., Zhao, Z., Zhang, M., Zeng, X.: Mc-slt: Towards low- resource signer-adaptive sign language translation. In: Proceed- ings of the 30th ACM International Conference on Multimedia, pp. 4939–4947 (2022)	Deaf involvement: 	no	Evaluation: there are no deaf people involved: only focus on the MT-process. Level -1	Level: 	-1	b	Depth: 	Only (hearing) researchers	Breadth: 	No different groups/ variations are involved	Scope: 	In none of the research life cycle steps
\item	Jin, T., Zhao, Z., Zhang, M., Zeng, X.: Prior knowledge and memory enriched transformer for sign language translation. In: Findings of the Association for Computational Linguistics: ACL 2022, pp. 3766–3775 (2022)	Deaf involvement: 	no	Evaluation: there are no deaf people involved: only focus on the MT-process. Level -1	Level: 	-1	b	Depth: 	Only (hearing) researchers	Breadth: 	No different groups/ variations are involved	Scope: 	In none of the research life cycle steps
\item	Kamata, K., Yoshida, T., Watanabe, M., \& Usui, Y. (1989). An approach to Japanese-sign language translation system. In Conference proceedings., IEEE international conference on systems, man and cybernetics (pp. 1089–1090). IEEE	Deaf involvement: 	no	Evaluation: there are no deaf people involved: only focus on the MT-process. Level -1	Level: 	-1	b	Depth: 	Only (hearing) researchers	Breadth: 	No different groups/ variations are involved	Scope: 	In none of the research life cycle steps
\item	Kan, J., Hu, K., Hagenbuchner, M., Tsoi, A.C., Bennamoun, M., Wang, Z.: Sign language translation with hierarchical spatio- temporal graph neural network. In: Proceedings of the IEEE/ CVF Winter Conference on Applications of Computer Vision, pp. 3367–3376 (2022)	Deaf involvement: 	no	Evaluation: there are no deaf people involved: only focus on the MT-process. Level -1	Level: 	-1	b	Depth: 	Only (hearing) researchers	Breadth: 	No different groups/ variations are involved	Scope: 	In none of the research life cycle steps
\item	Kim, S., Kim, C.J., Park, H.-M., Jeong, Y., Jang, J.Y., Jung, H.: Robust keypoint normalization method for korean sign language translation using transformer. In: 2020 International Conference on Information and Communication Technology Convergence (ICTC), pp. 1303–1305 (2020). \url{https://doi.org/10.1109/ICTC49870.2020.9289551}. IEEE	Deaf involvement: 	Yes, for training (16 signers) and testing of data (4)	Evaluation: there are no deaf people involved: only focus on the MT-process. Level -1	Level: 	-1	b	Depth: 	Only (hearing) researchers	Breadth: 	No different groups/ variations are involved	Scope: 	In none of the research life cycle steps
\item	Kouremenos, D., Ntalianis, K., \& Kollias, S. 2018. A novel rule based machine translation scheme from Greek to Greek sign language: Production of different types of large corpora and language models evaluation. 51, 110–135,	Deaf involvement: 	No	Evaluation: level -1. A translator, Human evaluation is fundamental and remains of crucial importance to proper assessment of the quality of MT systems. When the output of an MT system is evaluated, however, the whole process is taken into account. In our case, different aspects of the proposed RBMT system are evaluated such as: (a) all stages of development of the transfer rules, (b) accuracy of translation and (c) complexity.Thus, it cannot be understood by Deaf people, who cannot read the Greek language. A complete MT system for the GSL should produce animations, while a genuine and proper evaluation should involve Deaf peo?ple, measuring comprehension regarding the animated output 	Level: 	-1	b	Depth: 	Only (hearing) researchers	Breadth: 	No different groups/ variations are involved	Scope: 	In none of the research life cycle steps
\item	Kumar, S.S., Wangyal, T., Saboo, V., Srinath, R.: Time series  neural networks for real time sign language translation. In: 2018  17th IEEE International Conference on Machine Learning and  Applications (ICMLA), pp. 243–248 (2018). \url{https://doi.org/10.1109/ICMLA.2018.00043}. IEEE	Deaf involvement: 	no	Evaluation: there are no deaf people involved: only focus on the MT-process. Level -1	Level: 	-1	b	Depth: 	Only (hearing) researchers	Breadth: 	No different groups/ variations are involved	Scope: 	In none of the research life cycle steps
\item	Li, D., Xu, C., Yu, X., Zhang, K., Swift, B., Suominen, H., Li, H.: Tspnet: Hierarchical feature learning via temporal semantic pyramid for sign language translation. Adv. Neural. Inf. Process. Syst. 33, 12034–12045 (2020)	Deaf involvement: 	no	Evaluation: there are no deaf people involved: only focus on the MT-process. Level -1	Level: 	-1	b	Depth: 	Only (hearing) researchers	Breadth: 	No different groups/ variations are involved	Scope: 	In none of the research life cycle steps
\item	Li, R., Meng, L.: Sign language recognition and translation network based on multi-view data. Appl. Intell. 52(13), 14624– 14638 (2022)	Deaf involvement: 	no	Evaluation: there are no deaf people involved: only focus on the MT-process. Level -1	Level: 	-1	b	Depth: 	Only (hearing) researchers	Breadth: 	No different groups/ variations are involved	Scope: 	In none of the research life cycle steps
\item	López-Ludeña, V., San-Segundo, R., Morcillo, C. G., López, J. C., \& Muñoz, J. M. P. (2013). Increasing adaptability of a speech into sign language translation system. Expert Systems with Applications, 40(4), 1312–1322.	Deaf involvement: 	No	Evaluation: there are no deaf people involved: only focus on the MT-process. Level -1	Level: 	-1	b	Depth: 	Only (hearing) researchers	Breadth: 	No different groups/ variations are involved	Scope: 	In none of the research life cycle steps
\item	Luqman, H., Mahmoud, S.A.: A machine translation system from arabic sign language to arabic. Univ. Access Inf. Soc. 19(4), 891–904 (2020). \url{https://doi.org/10.1007/s10209-019-00695-6}	Deaf involvement: 	no	Evaluation: There are no deaf people involved: level -1. Evaluation by hearing Arab speakers for translation-evaluation (as the output is Arabic)	Level: 	-1	b	Depth: 	Only (hearing) researchers	Breadth: 	No different groups/ variations are involved	Scope: 	In none of the research life cycle steps
\item	Marshall, I., \& Sáfár, É. (2002). Sign language generation using HPSG. In Proceedings of the 9th Conference on Theoretical and Methodological Issues in Machine Translation of Natural Languages: Papers.	Deaf involvement: 	No	Evaluation: There are no deaf people involved (level -1) but the authors are aware of cco-creation: Sign research has frequently been carried out by hearingpeople usingdeaf informants and hence insights are typically second-hand. Additionally, the status of deaf informants themselves within the Deaf community raises a significant issue. Typically only 5-10\% of deaf people are born to deaf parents and thus are viewed as the genuine native signers who should act as informants and who should be asked to identify the preferred manner of signing a proposition rather than merely acceptable signing(Neidle et al. 2000). Deaf informants with hearingresearchers and initial review by hearingsigners are used to establish initial hypotheses. More extensive review by deaf users of the generated signing provides detailed feedback and guides revision.	Level: 	-1	b	Depth: 	Only (hearing) researchers	Breadth: 	No different groups/ variations are involved	Scope: 	In none of the research life cycle steps
\item	Marshall, I., \& Sáfár, É. (2003). A prototype text to British Sign Language (BSL) translation system. In The companion volume to the proceedings of 41st annual meeting of the association for computational linguistics (pp. 113–116).	Deaf involvement: 		Evaluation: there are no deaf people involved: only focus on the MT-process. Level -1	Level: 	-1	b	Depth: 	Only (hearing) researchers	Breadth: 	No different groups/ variations are involved	Scope: 	In none of the research life cycle steps
\item	Miranda, P.B., Casadei, V., Silva, E., Silva, J., Alves, M., Severo, M., Freitas, J.P.: Tspnet-hf: A hand/face tspnet method for sign language translation. In: Ibero-American Conference on Artifi- cial Intelligence, pp. 305–316 (2022). Springer	Deaf involvement: 	no	Evaluation: there are no deaf people involved: only focus on the MT-process. Level -1	Level: 	-1	b	Depth: 	Only (hearing) researchers	Breadth: 	No different groups/ variations are involved	Scope: 	In none of the research life cycle steps
\item	Mohamed, A., Hefny, H., et al.: A deep learning approach for gloss sign language translation using transformer. Journal of Computing and Communication 1(2), 1–8 (2022)	Deaf involvement: 	no	Evaluation: there are no deaf people involved: only focus on the MT-process. Level -1	Level: 	-1	b	Depth: 	Only (hearing) researchers	Breadth: 	No different groups/ variations are involved	Scope: 	In none of the research life cycle steps
\item	Morrissey, S. (2008). Assistive translation technology for deaf people: translating into and animating Irish sign language.	Deaf involvement: 		Evaluation: there are no deaf people involved: only focus on the MT-process. Level -1	Level: 	-1	b	Depth: 	Only (hearing) researchers	Breadth: 	Only (hearing) researchers	Scope: 	Only (hearing) researchers
\item	Morrissey, S., \& Way, A. (2005). An example-based approach to translating sign language.	Deaf involvement: 		Evaluation: there are no deaf people involved: only focus on the MT-process. Level -1	Level: 	-1	b	Depth: 	Only (hearing) researchers	Breadth: 	No different groups/ variations are involved	Scope: 	In none of the research life cycle steps
\item	Morrissey, S., \& Way, A. (2006). Lost in translation: the problems of using mainstream MT evaluation metrics for sign language translation	Deaf involvement: 	The authors mention: Clearly, in addition, human evaluation remains crucial for all such approaches.	Evaluation: there are no deaf people involved: only focus on the MT-process. Level -1	Level: 	-1	b	Depth: 	Only (hearing) researchers	Breadth: 	No different groups/ variations are involved	Scope: 	In none of the research life cycle steps
\item	Morrissey, S., Way, A., Stein, D., Bungeroth, J., Ney, H.: Com- bining data-driven mt systems for improved sign language trans- lation. In: European Association for Machine Translation (2007)	Deaf involvement: 	no	Evaluation: there are no deaf people involved: only focus on the MT-process. Level -1	Level: 	-1	b	Depth: 	Only (hearing) researchers	Breadth: 	No different groups/ variations are involved	Scope: 	In none of the research life cycle steps
\item	Moryossef, A., Yin, K., Neubig, G., Goldberg, Y.: Data aug- mentation for sign language gloss translation. In: Proceedings of the 1st International Workshop on Automatic Translation for Signed and Spoken Languages (AT4SSL), pp. 1–11. Association for Machine Translation in the Americas, Virtual (2021). \url{https://aclanthology.org/2021.mtsummit-at4ssl.1}	Deaf involvement: 	No	Evaluation: there are no deaf people involved: only focus on the MT-process. Level -1	Level: 	-1	b	Depth: 	Only (hearing) researchers	Breadth: 	No different groups/ variations are involved	Scope: 	In none of the research life cycle steps
\item	Nießen, S., \& Ney, H. (2004). Statistical machine translation with scarce resources using morpho-syntactic information. Computational Linguistics, 30(2), 181–204.	Deaf involvement: 	no	Evaluation: there are no deaf people involved: only focus on the MT-process. Level -1	Level: 	-1	b	Depth: 	Only (hearing) researchers	Breadth: 	No different groups/ variations are involved	Scope: 	In none of the research life cycle steps
\item	Orbay, A., Akarun, L.: Neural sign language translation by learn- ing tokenization. In: 2020 15th IEEE International Conference on Automatic Face and Gesture Recognition (FG 2020), pp. 222–228 (2020). IEEE	Deaf involvement: 	No	Evaluation: there are no deaf people involved: only focus on the MT-process. Level -1	Level: 	-1	b	Depth: 	Only (hearing) researchers	Breadth: 	No different groups/ variations are involved	Scope: 	In none of the research life cycle steps
\item	Othman, A., Jemni, M.: English-asl gloss parallel corpus 2012:  Aslg-pc12. In: 5th Workshop on the Representation and Pro?cessing of Sign Languages: Interactions Between Corpus and  Lexicon LREC (2012)	Deaf involvement: 	No	Evaluation: there are no deaf people involved: only focus on the MT-process. Level -1	Level: 	-1	b	Depth: 	Only (hearing) researchers	Breadth: 	No different groups/ variations are involved	Scope: 	In none of the research life cycle steps
\item	Partaourides, H., Voskou, A., Kosmopoulos, D., Chatzis, S., Metaxas, D.N.: Variational bayesian sequence-to-sequence net- works for memory-efficient sign language translation. In: Inter- national Symposium on Visual Computing, pp. 251–262 (2020). Springer	Deaf involvement: 	no	Evaluation: there are no deaf people involved: only focus on the MT-process. Level -1	Level: 	-1	b	Depth: 	Only (hearing) researchers	Breadth: 	No different groups/ variations are involved	Scope: 	In none of the research life cycle steps
\item	Porta, J., López-Colino, F., Tejedor, J., \& Colás, J. (2014). A rule-based translation from written Spanish to Spanish Sign Language glosses. Computer Speech and Language, 28(3), 788–811.	Deaf involvement: 	no	Evaluation: Level -1 A parallel Spanish-LSE corpus has bvbeen created by two hearing interpreters (one of them was CODA)	Level: 	-1	b	Depth: 	Only (hearing) researchers	Breadth: 	No different groups/ variations are involved	Scope: 	In none of the research life cycle steps
\item	Sáfár, É., \& Marshall, I. (2001). The architecture of an english-text-to-sign-languages translation system. In Recent advances in natural language processing RANLP, (pp. 223–228). Tzigov Chark Bulgaria.	Deaf involvement: 		Evaluation: there are no deaf people involved: only focus on the MT-process. Level -1	Level: 	-1	b	Depth: 	Only (hearing) researchers	Breadth: 	No different groups/ variations are involved	Scope: 	In none of the research life cycle steps
\item	Sáfár, É., \& Marshall, I. (2002). Sign Language Translation via DRT and HPSG. Conference on Intelligent Text Processing and Computational Linguistics.	Deaf involvement: 	No	Evaluation: there are no deaf people involved: only focus on the MT-process. Level -1	Level: 	-1	b	Depth: 	Only (hearing) researchers	Breadth: 	No different groups/ variations are involved	Scope: 	In none of the research life cycle steps
\item	San Segundo, R., Pérez, A., Ortiz, D., Luis Fernando, D., Torres, M. I., \& Casacuberta, F. (2007). Evaluation of alternatives on speech to sign language translation. In INTERSPEECH (pp. 2529–2532). Citeseer.	Deaf involvement: 	No	Evaluation: there are no deaf people involved: only focus on the MT-process. Level -1	Level: 	-1	b	Depth: 	Only (hearing) researchers	Breadth: 	No different groups/ variations are involved	Scope: 	In none of the research life cycle steps
\item	San-Segundo, R., Barra, R., Córdoba, R., D’Haro, L. F., Fernández, F., Ferreiros, J., et al. (2008). Speech to sign language translation system for Spanish. Speech Communication, 50(11–12), 1009–1020.	Deaf involvement: 	No	Evaluation: there are no deaf people involved: only focus on the MT-process. Level -1	Level: 	-1	b	Depth: 	Only (hearing) researchers	Breadth: 	No different groups/ variations are involved	Scope: 	In none of the research life cycle steps
\item	San-Segundo, R., Barra, R., D’Haro, L., Montero, J. M., Córdoba, R., \& Ferreiros, J. (2006). A spanish speech to sign language translation system for assisting deaf-mute people. In Ninth international conference on spoken language processing.	Deaf involvement: 	No	Evaluation: there are no deaf people involved: only focus on the MT-process. Level -1	Level: 	-1	b	Depth: 	Only (hearing) researchers	Breadth: 	No different groups/ variations are involved	Scope: 	In none of the research life cycle steps
\item	Saunders, B., Camgoz, N. C., \& Bowden, R. (2020b). Progressive transformers for end- to-end sign language production. In European conference on computer vision (pp. 687–705). Springer.	Deaf involvement: 	No	Evaluation: there are no deaf people involved: only focus on the MT-process. Level -1	Level: 	-1	b	Depth: 	Only (hearing) researchers	Breadth: 	No different groups/ variations are involved	Scope: 	In none of the research life cycle steps
\item	Schmidt, C., Koller, O., Ney, H., Hoyoux, T., Piater, J.: Using  viseme recognition to improve a sign language translation sys?tem. In: International Workshop on Spoken Language Transla?tion, pp. 197–203 (2013). Citeseer	Deaf involvement: 	No	Evaluation: there are no deaf people involved: only focus on the MT-process. Level -1	Level: 	-1	b	Depth: 	Only (hearing) researchers	Breadth: 	No different groups/ variations are involved	Scope: 	In none of the research life cycle steps
\item	Stein, D., Dreuw, P., Ney, H., Morrissey, S., Way, A.: Hand in hand: automatic sign language to English translation. In: Proceedings of the 11th Conference on Theoretical and Methodological Issues in Machine Translation of Natural Languages: Papers, Skövde, Sweden (2007). \url{https://aclanthology.org/2007.tmi-papers.26}	Deaf involvement: 	No	Evaluation: there are no deaf people involved: only focus on the MT-process. Level -1	Level: 	-1	b	Depth: 	Only (hearing) researchers	Breadth: 	No different groups/ variations are involved	Scope: 	In none of the research life cycle steps
\item	Stein, D., Schmidt, C., Ney, H.: Analysis, preparation, and opti?mization of statistical sign language machine translation. Mach.  Transl. 26(4), 325–357 (2012)	Deaf involvement: 		Evaluation: there are no deaf people involved: only focus on the MT-process. Level -1	Level: 	-1	b	Depth: 	Only (hearing) researchers	Breadth: 	No different groups/ variations are involved	Scope: 	In none of the research life cycle steps
\item	Stein, D., Schmidt, C., Ney, H.: Sign language machine transla- tion overkill. In: International Workshop on Spoken Language Translation (IWSLT) 2010 (2010)	Deaf involvement: 	no	Evaluation: there are no deaf people involved: only focus on the MT-process. Level -1	Level: 	-1	b	Depth: 	Only (hearing) researchers	Breadth: 	No different groups/ variations are involved	Scope: 	In none of the research life cycle steps
\item	Stoll, S., Camgöz, N. C., Hadfield, S., \& Bowden, R. (2018). Sign language production us- ing neural machine translation and generative adversarial networks. In Proceedings of the 29th British machine vision conference (BMVC 2018). University of Surrey.	Deaf involvement: 	no	Evaluation: there are no deaf people involved: only focus on the MT-process. Level -1	Level: 	-1	b	Depth: 	Only (hearing) researchers	Breadth: 	No different groups/ variations are involved	Scope: 	In none of the research life cycle steps
\item	Tarres, L., Gállego, G.I., Giro-i-Nieto, X., Torres, J.: Tackling low-resourced sign language translation: Upc at wmt-slt 22. In: Proceedings of the Seventh Conference on Machine Transla- tion, pp. 994–1000. Association for Computational Linguistics, Abu Dhabi (2022). \url{https://aclanthology.org/2022.wmt-1.97}	Deaf involvement: 	no	Evaluation: there are no deaf people involved: only focus on the MT-process. Level -1	Level: 	-1	b	Depth: 	Only (hearing) researchers	Breadth: 	No different groups/ variations are involved	Scope: 	In none of the research life cycle steps
\item	Tokuda, M., \& Okumura, M. (1998). Towards automatic translation from japanese into japanese sign language. In Assistive technology and artificial intelligence (pp. 97–108). Springer.	Deaf involvement: 	No	Evaluation: there are no deaf people involved: only focus on the MT-process. Level -1	Level: 	-1	b	Depth: 	Only (hearing) researchers	Breadth: 	No different groups/ variations are involved	Scope: 	In none of the research life cycle steps
\item	Wazalwar, S. S., \& Shrawankar, U. (2017). Interpretation of sign language into English using NLP techniques. Journal of Information and Optimization Sciences, 38(6), 895–910.	Deaf involvement: 	no	Evaluation: The videos were interpretered by hearing teachers of school for the deaf Level -1	Level: 	-1	a	Depth: 	Only (hearing) researchers	Breadth: 	No different groups/ variations are involved	Scope: 	In none of the research life cycle steps
\item	Yin, A., Zhao, Z., Jin, W., Zhang, M., Zeng, X., He, X.: Mlslt: Towards multilingual sign language translation. In: Proceed- ings of the IEEE/CVF Conference on Computer Vision and Pattern Recognition, pp. 5109–5119 (2022)	Deaf involvement: 	no	Evaluation: there are no deaf people involved: only focus on the MT-process. Level -1	Level: 	-1	b	Depth: 	Only (hearing) researchers	Breadth: 	No different groups/ variations are involved	Scope: 	In none of the research life cycle steps
\item	Yin, A., Zhao, Z., Liu, J., Jin, W., Zhang, M., Zeng, X., He, X.: Simulslt: End-to-end simultaneous sign language translation. In: Proceedings of the 29th ACM International Conference on Mul- timedia, pp. 4118–4127 (2021)	Deaf involvement: 	no	Evaluation: there are no deaf people involved: only focus on the MT-process. Level -1	Level: 	-1	b	Depth: 	Only (hearing) researchers	Breadth: 	No different groups/ variations are involved	Scope: 	In none of the research life cycle steps
\item	Yin, K., Read, J.: Better sign language translation with stmc- transformer. In: Proceedings of the 28th International Conference on Computational Linguistics, pp. 5975–5989 (2020). \url{https://doi.org/10.18653/v1/2020.coling-main.525}	Deaf involvement: 	no	Evaluation: there are no deaf people involved: only focus on the MT-process. Level -1	Level: 	-1	b	Depth: 	Only (hearing) researchers	Breadth: 	No different groups/ variations are involved	Scope: 	In none of the research life cycle steps
\item	Zhang, X., Duh, K.: Approaching sign language gloss translation as a low-resource machine translation task. In: Proceedings of the 1st International Workshop on Automatic Translation for Signed and Spoken Languages (AT4SSL), pp. 60–70. Association for Machine Translation in the Americas, Virtual (2021). \url{https://aclantholo gy.org/2021.mtsummit-at4ssl.7}	Deaf involvement: 	no	Evaluation: there are no deaf people involved: only focus on the MT-process. Level -1	Level: 	-1	b	Depth: 	Only (hearing) researchers	Breadth: 	No different groups/ variations are involved	Scope: 	In none of the research life cycle steps
\item	Zhao, J., Qi, W., Zhou, W., Duan, N., Zhou, M., \& Li, H. (2021). Conditional sentence generation and cross-modal reranking for sign language translation. IEEE Transactions on Multimedia, 24, 2662–2672.	Deaf involvement: 	no	Evaluation: there are no deaf people involved: only focus on the MT-process. Level -1	Level: 	-1	b	Depth: 	Only (hearing) researchers	Breadth: 	No different groups/ variations are involved	Scope: 	In none of the research life cycle steps
\item	Zhao, L., Kipper, K., Schuler, W., Vogler, C., Badler, N., \& Palmer, M. (2000). A machine translation system from English to American sign language. In Conference of the association for machine translation in the Americas (pp. 54–67). Springer.	Deaf involvement: 	no	Evaluation: there are no deaf people involved: only focus on the MT-process. Level -1	Level: 	-1	b	Depth: 	Only (hearing) researchers	Breadth: 	No different groups/ variations are involved	Scope: 	In none of the research life cycle steps
\item	Zheng, J., Chen, Y., Wu, C., Shi, X., \& Kamal, S. M. (2021). Enhancing neural sign lan- guage translation by highlighting the facial expression information. Neurocomputing, 464, 462–472.	Deaf involvement: 	No	Evaluation: there are no deaf people involved: only focus on the MT-process. Level -1	Level: 	-1	b	Depth: 	Only (hearing) researchers	Breadth: 	No different groups/ variations are involved	Scope: 	In none of the research life cycle steps
\item	Zheng, J., Zhao, Z., Chen, M., Chen, J., Wu, C., Chen, Y., Shi, X., Tong, Y.: An improved sign language translation model with explainable adaptations for processing long sign sentences. Com- putational Intelligence and Neuroscience 2020 (2020)	Deaf involvement: 	no	Evaluation: there are no deaf people involved: only focus on the MT-process. Level -1	Level: 	-1	b	Depth: 	Only (hearing) researchers	Breadth: 	No different groups/ variations are involved	Scope: 	In none of the research life cycle steps
\item	Zhou, H., Zhou, W., Zhou, Y., Li, H.: Spatial-temporal multi- cue network for sign language recognition and translation. IEEE Trans. Multimedia (2021). \url{https://doi.org/10.1109/TMM.2021.3059098}	Deaf involvement: 	no	Evaluation: there are no deaf people involved: only focus on the MT-process. Level -1	Level: 	-1	b	Depth: 	Only (hearing) researchers	Breadth: 	No different groups/ variations are involved	Scope: 	In none of the research life cycle steps
\item	Baldassarri, S., Cerezo, E., \& Royo-Santas, F. (2009). Automatic translation sys- tem to spanish sign language with a virtual interpreter. In IFIP conference on human-computer interaction (pp. 196–199). Springer.	Deaf involvement: 	Yes, two teachers of a school for interpreters	Evaluation: Level 0? Or level 1? Assessment was done by two teachers of a school of interpreters considering the  accuracy of two aspects: the translation and the synthesis of the signs by the virtual  interpreter.	Level: 	0		Depth: 	(hearing) researchers, SL-interpreters, but not the SL-user	Breadth: 	little variation (not the SL-user involved)	Scope: 	In the evaluation/ reflection phrase
\item	Camgoz, N.C., Hadfield, S., Koller, O., Ney, H., Bowden, R.: Neural sign language translation. In: Proceedings of the IEEE Conference on Computer Vision and Pattern Recognition, pp. 7784–7793 (2018). \url{https://doi.org/10.1109/CVPR.2018.00812}	Deaf involvement: 	no	Evaluation: Level 0. For the corpus, they used 9 different signers. Furthermore, the corpus annotations are made by SL-interpreters and deaf specialists.	Level: 	0		Depth: 		Breadth: 		Scope: 	
\item	Camgöz, N.C., Saunders, B., Rochette, G., Giovanelli, M., Inches, G., Nachtrab-Ribback, R., Bowden, R.: Content4all open research sign language translation datasets. In: 2021 16th IEEE International Conference on Automatic Face and Gesture Recog- nition (FG 2021), pp. 1–5 (2021). \url{https://doi.org/10.1109/FG52635.2021.9667087}	Deaf involvement: 		Evaluation: Level 0. There were deaf experts and SL interpreters for the match ing of the SpLs text with the corresponding SL-video pairs, and annotation process	Level: 	0		Depth: 		Breadth: 		Scope: 	
\item	Dal Bianco, P., Ríos, G., Ronchetti, F., Quiroga, F., Stanchi, O., Hasperué, W., Rosete, A.: Lsa-t: The first continuous argentinian sign language dataset for sign language translation. In: Ibero- American Conference on Artificial Intelligence, pp. 293–304 (2022). Springer	Deaf involvement: 	no, a generated corpus from videos of YouTube	Evaluation: Level 0. Videos of channel CN Sordos, a news channel created by deaf people and deaf people's relatives. 103 deaf signers as guests	Level: 	0		Depth: 		Breadth: 		Scope: 	
\item	Ebling, S., \& Huenerfauth, M. (2015). Bridging the gap between sign language machine translation and sign language animation using sequence classification. In Proceedings of SLPAT 2015: 6th workshop on speech and language processing for assistive technologies (pp. 2–9).	Deaf involvement: 	Yes, deaf and hearing team members (translating)	Evaluation: Yes, deaf and hearing team members (translating), but not clear to what exten (level 0)	Level: 	0		Depth: 		Breadth: 		Scope: 	
\item	Ko, S.-K., Kim, C.J., Jung, H., Cho, C.: Neural sign language  translation based on human keypoint estimation. Appl. Sci.  9(13), 2683 (2019)	Deaf involvement: 	no	Evaluation: level 0. 14 hearing-impaired for recordings (a copy of an 'expert' signing the requested signs, which the signers needed to copy)	Level: 	0		Depth: 		Breadth: 		Scope: 	
\item	Kr\v{n}oul, Z., Kanis, J., \v{Z}elezny, M., \& M\"uller, L. (2007). Czech text-to-sign speech ` synthesizer. In International workshop on machine learning for multimodal interaction (pp. 180–191). Springer.	Deaf involvement: 	two participants for the evaluation of the Sign Speech synthesizer	Evaluation: Level 0. two experts in SignSpeech for the evaluation of the Sign Speech synthesizer	Level: 	0		Depth: 		Breadth: 		Scope: 	
\item	Mass\'o, G., \& Badia, T. (2010). Dealing with sign language morphemes in statistical machine translation. In 4th workshop on the representation and processing of sign languages: Corpora and sign language technologies, Valletta, Malta (pp. 154–157). Matthes, S., Hanke, T., Regen, A., Storz, J., Worseck, S., Efthimiou, E., et al. (2012).	Deaf involvement: 	No, the authors mention: Unfortunately, it was not possible to conduct a human evaluation by native Deaf signers	Evaluation: (level 0) The authors created a corpus based on Catalan Weather texts which were translated by a native Deaf signers	Level: 	0		Depth: 		Breadth: 		Scope: 	
\item	Moe, S.Z., Thu, Y.K., Thant, H.A., Min, N.W., Supnithi, T.: Unsupervised Neural Machine Translation between Myanmar Sign Language and Myanmar Language . TIC 14(15), 16 (2020)	Deaf involvement: 	Yes, for data collection	Evaluation: Level 0. 30 SL trainers and deaf people from different MSL dialects for data collection	Level: 	0		Depth: 		Breadth: 		Scope: 	
\item	Moe, S.Z., Thu, Y.K., Thant, H.A., Min, N.W.: Neural Machine Translation between Myanmar Sign Language and Myanmar Written Text. In: the Second Regional Conference on Optical Character Recognition and Natural Language Processing Technologies  for ASEAN Languages, pp. 13–14 (2018)	Deaf involvement: 	no	Evaluation: Level 0. Yes, data collection of 22 SL-trainers, and deaf deople with different MSL dialects and different ages	Level: 	0		Depth: 		Breadth: 		Scope: 	
\item	Morrissey, S. (2011). Assessing three representation methods for sign language machine translation and evaluation. In Proceedings of the 15th annual meeting of the European association for machine translation (EAMT 2011), Leuven, Belgium (pp. 137–144). Citeseer.	Deaf involvement: 	However, the authors pointed out that, given the auto?matic evaluation used, it was not clear which was the best format and that experiments should be accompanied by human evaluation to ascertain the translation quality	Evaluation: Level 0. A native ISL signer manually translated and signed the dialogue in ISL	Level: 	0		Depth: 		Breadth: 		Scope: 	
\item	Müller, M., Ebling, S., Avramidis, E., Battisti, A., Berger, M.,  Bowden, R., Brafort, A., Cihan Camgöz, N., España-Bonet, C.,  Grundkiewicz, R., Jiang, Z., Koller, O., Moryossef, A., Perrollaz,  R., Reinhard, S., Rios, A., Shterionov, D., Sidler-Miserez, S.,  Tissi, K., Van Landuyt, D.: Findings of the frst wmt shared task  on sign language translation (wmt-slt22). In: Proceedings of the  Seventh Conference on Machine Translation, pp. 744–772. Asso?ciation for Computational Linguistics, Abu Dhabi (2022). \url{https://aclanthology.org/2022.wmt-1.71}	Deaf involvement: 	Seven teams participated, four native German speakers who were educated interpreters	Evaluation: Level 0. Manually correction of subtitles by deaf signers, evaluators trained DGSG interpreters	Level: 	0		Depth: 		Breadth: 		Scope: 	
\item	Rodriguez, J., Martinez, F.: How important is motion in sign lan- guage translation? IET Comput. Vision 15(3), 224–234 (2021)	Deaf involvement: 	No	Evaluation: Level 0. 9 deaf signers and 2 CODAs for the recordings of the dataset	Level: 	0		Depth: 		Breadth: 		Scope: 	
\item	Al-Khalifa, H. S. (2010). Introducing Arabic sign language for mobile phones. In International conference on computers for handicapped persons (pp. 213–220). Springer.	Deaf involvement: 	Evaluating of the system, not clear if the group of users were deaf	Evaluation: Five deaf (3) and non-deaf (2)people answered a survey (level 1)	Level: 	1		Depth: 		Breadth: 		Scope: 	
\item	Chiu, Y.-H., Wu, C.-H., Su, H.-Y., \& Cheng, C.-J. (2006). Joint optimization of word alignment and epenthesis generation for Chinese to Taiwanese sign synthesis. IEEE Transactions on Pattern Analysis and Machine Intelligence, 29(1), 28–39.	Deaf involvement: 	Subjective evaluation (with missing how many, who are the subjects etc)	Evaluation: Level 1 Five profoundly deaf students in the sixth grade evaluatd the utility of the proposed approach as practical learning aid	Level: 	1		Depth: 	5 SL-users + (hearing) researchers	Breadth: 	Litlle variation	Scope: 	In the evaluation/ reflection phrase
\item	Hilzensauer, M., Krammer, K.: A multilingual dictionary for  sign languages:``spreadthesign''. ICERI2015 Proceedings,  7826–7834 (2015)	Deaf involvement: 	no	Evaluation: Level 1. Fifteen partner countries, according to a list, which were discussed with deaf collaborators - who then chose the signs / or sign dialects	Level: 	1		Depth: 		Breadth: 		Scope: 	
\item	Khan, N. S., Abid, A., \& Abid, K. (2020). A novel natural language processing (NLP)– based machine translation model for English to Pakistan sign language translation. Cognitive Computation, 12, 748–765.	Deaf involvement: 	Yes, deaf scholars for evaluation (amount is not mentioned)	Evaluation: Level 1, Translating English sentences into PSL sentences with the help of SL interpreters and three deaf subjects for recordings, also used for evaluation.	Level: 	1		Depth: 		Breadth: 		Scope: 	
\item	López-Ludeña, V., San-Segundo, R., Montero, J. M., Córdoba, R., Ferreiros, J., \& Pardo, J. M. (2012). Automatic categorization for improving Spanish into Spanish Sign Language machine translation. Computer Speech and Language, 26(3), 149–167.	Deaf involvement: 	Two experts in LSE, who were also involved into the corpus generation, but the authors aknowledged that deaf people also should be evaluate how the avatar represents these signs	Evaluation: Level 1:"These sentences were translated into LSE, both in text (sequence of signs) and in video, and compiled in an excel file. The translation was carried out by two LSE experts in parallel. When there was any discrepancy between them, a committee of four people (one Spanish linguist, 2 deaf LSE experts, and a Spanish linguistic expert on LSE) who knew LSE took the decision: select one of the LSE expert proposals, propose a new one translation alternative, or considering both proposals as alternative translations.	Level: 	1		Depth: 	Two LSE experts (for translation) , Spanish linguist, 2 deaf LSE experts and a Spanish- LSE experts	Breadth: 	little variation but the SL-user is involved	Scope: 	implementation, reflection
\item	Luqman, H., \& Mahmoud, S. A. (2019). Automatic translation of Arabic text-to-Arabic sign language. Universal Access in the Information Society, 18(4), 939–951.	Deaf involvement: 	Yes, evaluation by 1 deaf person and 1 translator	Evaluation: level 1: based on wordlist 2 native signers for translating Arabic into ArSL, evaluation by 1 deaf person and 1 expert bilingual translator	Level: 	1		Depth: 	(hearing) researchers, SL-interpreters, three deaf subjects	Breadth: 	Little variation in the groups	Scope: 	implementation, reflection
\item	Rodriguez, J., Chacon, J., Rangel, E., Guayacan, L., Hernandez, C., Hernandez, L., Martinez, F.: Understanding motion in sign language: A new structured translation dataset. In: Proceedings of the Asian Conference on Computer Vision (2020)	Deaf involvement: 	Yes, for training and testing of data	Evaluation: Level 1. Five deaf signers out of different regios has been recorded, 10 signers for training and testing evaluation.	Level: 	1		Depth: 		Breadth: 		Scope: 	
\item	Sagawa, H., Ohki, M., Sakiyama, T., Oohira, E., Ikeda, H., \& Fujisawa, H. (1996). Pattern recognition and synthesis for a sign language translation system. Journal of Visual Languages and Computing, 7(1), 109–127.	Deaf involvement: 	Yes, 1 deaf person for data-collection (level -1 or level 0)	Evaluation: Four hearing-impaired and two interpreters evaluated the SL sentences (level 1)	Level: 	1		Depth: 	Only (hearing) researchers	Breadth: 	1 deaf person, four HoH persons, two interpreters	Scope: 	In data-collection and evaluation
\item	San-Segundo, R., López, V., Martın, R., Sánchez, D., Garcıa, A.: Language resources for spanish–spanish sign language (lse) translation. In: Proceedings of the 4th Workshop on the Repre- sentation and Processing of Sign Languages: Corpora and Sign Language Technologies at LREC, pp. 208–211 (2010)	Deaf involvement: 	Yes, ten deaf signers tested the system in a real-life sitation	Evaluation: Level 1. the first day was an information day about the project and the evaluation, the second day within 6 different scenarios was tested.	Level: 	1		Depth: 	Only (hearing) researchers	Breadth: 	No different groups/ variations are involved	Scope: 	In none of the research life cycle steps
\item	Stein, D., Bungeroth, J., \& Ney, H. (2006). Morpho-syntax based statistical methods for automatic sign language translation. In Proceedings of the 11th annual conference of the European association for machine translation.	Deaf involvement: 	Yes, for evaluation (2 deaf people)	Evaluation: Yes. For the rating of the coherence of a German sentence to the avatar outpunt (level 1)	Level: 	1		Depth: 	2 SL-users + (hearing) researchers	Breadth: 	Little variation	Scope: 	In evaluation/ reflection phrase
\item	Su, H.-Y., \& Wu, C.-H. (2009). Improving structural statistical machine translation for sign language with small corpus using thematic role templates as translation memory. IEEE Transactions on Audio, Speech, and Language Processing, 17(7), 1305–1315.	Deaf involvement: 	Yes, 10 dove studenten (divided into control and testgroup)	Evaluation: Level 1? The developed parallel bilingual corpus hInas been annotated and verified by 3 TSL linguists	Level: 	1		Depth: 	10 deaf students divided over 2 groups, and 3 TSL linguists	Breadth: 	Variation by two controle and testgroups, check by TSL linguists	Scope: 	implementation, reflection
\item	Wu, C.-H., Su, H.-Y., Chiu, Y.-H., \& Lin, C.-H. (2007). Transfer-based statistical translation of Taiwanese sign language using PCFG. ACM Transactions on Asian Language Information Processing (TALIP), 6(1), 1–es.	Deaf involvement: 	Subjective evaluation	Evaluation: Level 1: group 1: 10 hearing people who used TSL for years, group 2: 10 native TSL signers evaluated the translated sentences	Level: 	1		Depth: 	10 hearing people who used TSL for years + 10 native TSL signers + (hearing) researchers	Breadth: 	Variation by two groups (native and non-native signers	Scope: 	In the evaluation/ reflection phrase
\item	Zhou, H., Zhou, W., Qi, W., Pu, J., \& Li, H. (2021). Improving sign language translation with monolingual data by sign back-translation. In Proceedings of the IEEE/CVF conference on computer vision and pattern recognition (pp. 1316–1325).	Deaf involvement: 	no	Evaluation: level 1. SL linguistic experts, several SL teachers for design of the specific content, 10 native signers for video recording	Level: 	1		Depth: 		Breadth: 		Scope: 	
\item	Jantunen, T., Rousi, R., Rainò, P., Turunen, M., Moeen Valipoor, M., \& García, N. (2021). Is there any hope for developing automated translation technology for sign languages. Multilingual Facilitation, 61–73.	Deaf involvement: 		Evaluation: Level 2. There are NADs included, and also a paragraph about Co-Engineering, Participation and Culture	Level: 	2		Depth: 		Breadth: 		Scope: 	
\item	Morrissey, S., \& Way, A. (2007). Joining hands: Developing a sign language machine translation system with and for the deaf community.	Deaf involvement: 	two deaf signers for translation work anhd cosuiltation work + data-collection	Evaluation: Level 2? Yes, the involvement of Deaf colleagues, members of the Deaf community within the choice of a domain for SLT (by asking the Centre for Deaf Studies), the human translation, advice on the SL grammar and linguistics, manual evaluators of the translated output	Level: 	2		Depth: 	deaf collegeagues + (hearing) researchers	Breadth: 	Deaf studies, deaf colleagues (in team) and SLC	Scope: 	Initiation, planning, implementation, reflection

}
\end{enumerate}

\begin{adjustwidth}{-\extralength}{0cm}

\reftitle{References}


\bibliography{main}


%


\PublishersNote{}
\end{adjustwidth}
\end{document}